%% file: main.tex
\documentclass[11pt]{article}
\usepackage{makecell}

\usepackage{xcolor}
\usepackage{showexpl}
\usepackage{colortbl}
\usepackage[most]{tcolorbox}

\definecolor{orange}{HTML}{FF7F0E}
\definecolor{green}{HTML}{2CA02C}
\definecolor{darkred}{HTML}{cc0808}

\definecolor{main}{HTML}{5989cf}    
\definecolor{sub}{HTML}{cde4ff}     
\colorlet{LightOrange}{orange!25!}
\colorlet{Lightgreen}{green!25!}

\tcbset{
    orangeboxstyle/.style={
        boxsep=1.5pt, 
        left=2pt, right=2pt, 
        top=1pt, bottom=1pt, 
        colframe=white, 
        colback=LightOrange, 
        box align=base, 
        enhanced,
        nobeforeafter 
    },
    greenboxstyle/.style={
        boxsep=1.5pt, 
        left=2pt, right=2pt, 
        top=1pt, bottom=1pt, 
        colframe=white, 
        colback=Lightgreen, 
        box align=base, 
        enhanced,
        nobeforeafter 
    }
}

\tcbset{
    sharp corners,
    colback = white,
    before skip = 0.1cm,    
    after skip = 0.5cm      
}           

\newtcolorbox{boxH}{
    colback = sub, 
    colframe = main, 
    boxrule = 0pt, 
    leftrule = 5pt 
}

\newtcbox{\greenbox}{greenboxstyle}
\newtcbox{\orangebox}{orangeboxstyle}

\makeatletter
\newcommand\notsotiny{\@setfontsize\notsotiny{8}{9}}
\makeatother

\usepackage{authblk}
\usepackage[]{acl}

\usepackage{float}
\usepackage{graphicx}
\usepackage{multicol}
\usepackage{soul}

\usepackage{tcolorbox}
\usepackage{amsmath}
\usepackage{newtxmath}
\usepackage[normalem]{ulem}

\usepackage{times}
\usepackage{latexsym}
\usepackage{enumitem}
\usepackage{multirow}
\usepackage{booktabs}
\usepackage{tabularx}
\usepackage{longtable}
\usepackage{array}

\usepackage{arydshln}

\usepackage[T1]{fontenc}

\usepackage[utf8]{inputenc}

\usepackage{siunitx}
\usepackage{array}
\usepackage{microtype}
\usepackage{subcaption}

\usepackage{cleveref}
\crefformat{section}{\S#2#1#3}

\usepackage{booktabs}

\usepackage{balance}

\usepackage{graphicx}
\usepackage[frozencache=true,cachedir=minted-cache]{minted} 
\usepackage[inkscapearea=drawing]{svg}
\usepackage{multirow}
\usepackage{pbox}
\usepackage{xspace}
\usepackage{stfloats}

\usepackage{url}
\usepackage{tikz}
\usetikzlibrary{shapes, arrows, positioning}
\usepackage{twemojis}
\usepackage[breaklinks]{hyperref}

\svgpath{{./figures/}}
\usepackage{wrapfig}

\title{\Large\texttwemoji{bullseye} Localizing and Mitigating Errors in Long-form Question Answering}
\author{
\begin{minipage}[t]{\textwidth}
\centering
\normalsize
\bf

Rachneet Sachdeva$^{\spadesuit}$, Yixiao Song$^\heartsuit$, Mohit Iyyer$^\heartsuit$,
Iryna Gurevych$^{\spadesuit}$ \\
{\footnotesize \normalfont 
$^{\spadesuit}$Ubiquitous Knowledge Processing Lab (UKP Lab), \\Department of Computer Science and Hessian Center for AI (hessian.AI), \\Technical University of Darmstadt \\
$^\heartsuit$University of Massachusetts Amherst \\
\url{www.ukp.tu-darmstadt.de}
} 
\end{minipage}
}
\begin{document}
\maketitle

\begin{abstract}

\input{sections/1-abstract}
\end{abstract}

\section{Introduction}
\input{sections/2-introduction}

\section{Related Work}
\input{sections/3-related_work}

\section{HaluQuestQA~(HQ$^2$A)}
\input{sections/4-data_collection}
\label{sec: data_collection}

\section{Error Mitigation}

\input{sections/5-hallucination}

\section{Results}
\input{sections/6-results}

\section{Conclusion}
\input{sections/7-conclusion}

\label{sec:bibtex}

\section*{Limitations}
\input{sections/8-limitations}

\section*{Ethics and Broader Impact Statement}
\input{sections/9-ethics}

\section*{Acknowledgements}
\input{sections/acknowledgements}

\bibliography{main}
\bibliographystyle{acl_natbib}

\appendix
\clearpage
\input{sections/10-appendix}

\end{document}

%% file: sections/1-abstract.tex
Long-form question answering~(LFQA) aims to provide thorough and in-depth answers to complex questions, enhancing comprehension.
However, such detailed responses are prone to hallucinations and factual inconsistencies, challenging their faithful evaluation. 
This work introduces \textit{HaluQuestQA}, the first hallucination dataset with localized error annotations for human-written and model-generated LFQA answers.
HaluQuestQA comprises 698 QA pairs with 1.8k span-level error annotations for five different error types by expert annotators, along with preference judgments.
Using our collected data, we thoroughly analyze the shortcomings of long-form answers and find that they lack comprehensiveness and provide unhelpful references. 
We train an automatic feedback model on this dataset that predicts error spans with incomplete information and provides associated explanations.
Finally, we propose a prompt-based approach, \textit{Error-Informed Refinement}, that uses signals from the learned feedback model to refine generated answers, which we show reduces errors and improves the quality of the answers across multiple models. Furthermore, humans find the answers generated by our approach comprehensive and highly prefer them ($84\%$) over the baseline answers.\footnote{Code and data available at: \href{https://github.com/UKPLab/acl2025-lfqa-hallucination}{github.com/lfqa-errors}} 

%% file: sections/2-introduction.tex
 \begin{figure}[!t]
  \center
  \includegraphics[width=0.48\textwidth, height=6.7cm]{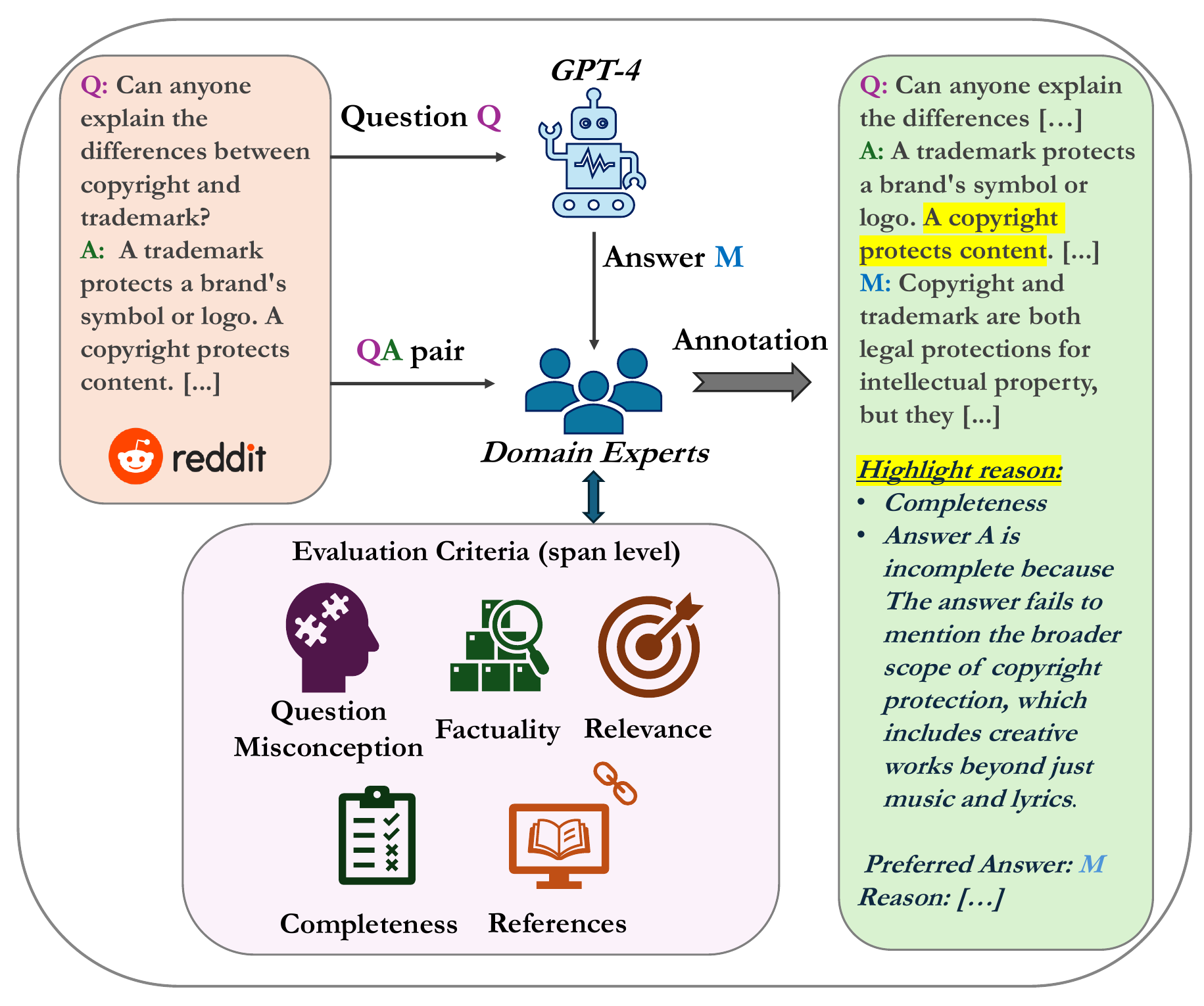}
    \caption{Overview of our data collection process. Using five fine-grained evaluation criteria, we collect \textit{span-level} expert human judgments on question-answer pairs from the Reddit platform, as well as on corresponding answers generated by GPT-4.}
  \label{fig:lfqa_eval}
  \vspace{-0.2cm}
  \end{figure}

Long-form question answering (LFQA) provides comprehensive, user-friendly, and in-depth responses to complex questions by leveraging state-of-the-art large language models~(LLMs) and retriever components ~\cite{krishna-etal-2021-hurdles, DBLP:journals/corr/abs-2112-09332}. 
While LLMs generate plausible and convincing answers, they also hallucinate and produce factually inconsistent, irrelevant, and incomplete content~\cite{DBLP:journals/corr/abs-2010-05478, laban-etal-2022-summac, DBLP:journals/corr/abs-2203-11147, DBLP:journals/corr/abs-2202-03629}, 
which are difficult to detect for both humans and machines.

Traditional evaluation metrics of answer quality, such as BLEU~\cite{papineni-etal-2002-bleu}, ROUGE~\cite{lin-2004-rouge}, and BERTScore~\cite{DBLP:conf/iclr/ZhangKWWA20} yield only a single score, obscuring the error \textit{type}, \textit{severity}, and \textit{location} in the answer. 
We take inspiration from machine translation, which moved beyond this simplistic evaluation paradigm by localizing and categorizing errors~\cite{DBLP:journals/corr/abs-2104-14478, DBLP:journals/corr/abs-2406-11580}, resulting in higher-quality and more interpretable evaluations. We make a similar contribution to the field of LFQA by asking human annotators to identify spans from the answers that correspond to errors and categorize each span into an error schema that we design. Our work is the first to explore error localization in LFQA, offering a more detailed and interpretable evaluation of answers.

LLMs make many errors for LFQA that require targeted evaluation. \citet{xu-etal-2023-critical} highlight that key aspects such as \textit{factuality, relevance, completeness, structure, references, and accessibility} are essential to evaluate long-form answers. Other recent studies have also validated the importance of aspects such as question misconception~\cite{krishna-etal-2021-hurdles}, factuality~\cite{DBLP:conf/nips/WuHSDSASOH23, jiang2023tigerscore, wang-etal-2024-revisiting}, relevance~\cite{tang2024tofueval}, and completeness~\cite{DBLP:journals/corr/abs-2501-03545} in determining answer quality. 
While prior work has mainly focused on evaluating factuality~\citep{lee2022factuality, DBLP:journals/corr/abs-2305-14251, DBLP:journals/corr/abs-2305-11747, DBLP:journals/corr/abs-2307-06908} and faithfulness~\cite{su-etal-2022-read} in long-form text generation, other aspects of evaluation, such as response completeness and relevance (which can particularly mislead users), have received less attention.

Our work addresses this gap by introducing \textit{HaluQuestQA}, a dataset of long-form answers annotated at the span level with five error types: \textit{question misconception, factuality, completeness, relevance, and references}. Expert annotators provide these annotations and preference judgments, as shown in \Cref{fig:lfqa_eval}.

Next, we train an automatic feedback model on this dataset to predict erroneous answer spans that lack key details to address the question comprehensively. 
The feedback model provides fine-grained feedback, identifying error locations (sentence level), error justification, and a confidence score, all without relying on reference texts~\cite{DBLP:conf/emnlp/XuWPSFWL23}.
Finally, we propose \textsc{Error-informed refinement}, a prompt-based approach that uses signals from the feedback model to refine generated answers~\cite{DBLP:conf/nips/MadaanTGHGW0DPY23}, reducing errors and improving answer quality across multiple LLMs.

\noindent Our contributions are summarized as follows:
\begin{itemize}[itemsep=0.4em, topsep=1pt, parsep=0pt, leftmargin=*]
\item  We release \textit{HaluQuestQA}, a dataset with span-level error annotations on pairs of human-written and model-generated answers. Our analysis reveals that long-form answers often lack comprehensiveness and provide unhelpful references.

\item We train a feedback model to identify erroneous answer spans with \textit{incomplete information}, aligned with expert human judgments. Although our dataset encompasses multiple errors, our feedback model focuses on completeness errors, which are identified as the most critical issue in the LFQA answers.

\item We propose Error-Informed Refinement, an approach that applies fine-grained feedback from our learned model to improve the quality of human-written and LLM-generated answers. 

\end{itemize}

%% file: sections/3-related_work.tex
\vspace{-0.1cm}
\paragraph{Human evaluation.}
Prior work~\cite{krishna-etal-2021-hurdles} has shown that human evaluation for LFQA tasks is challenging due to long answer lengths, and expert annotators are required to evaluate them effectively. 
\citet{xu-etal-2023-critical} hire expert annotators and identify nine multi-faceted aspects for meaningful LFQA evaluation. 
While some of these fine-grained aspects, such as factuality~\cite{DBLP:journals/corr/abs-2010-05478, laban-etal-2022-summac}, coherence~\cite{goyal-etal-2022-snac}, and completeness~\cite{tang2024tofueval}, have been used to investigate errors in summarization tasks, ours is amongst the first works to study LFQA-centric errors at the span level. 
To this end, we collect span-level annotations of LFQA errors, enabling high-quality and interpretable evaluations that can be used to improve answer quality. 
While this has been done for machine translation~\cite{DBLP:journals/corr/abs-2104-14478, DBLP:journals/corr/abs-2406-11580}, it has not yet been applied to long-form question answering.

\vspace{-0.2cm}
\paragraph{Automatic evaluation.}
Increasing focus on the reliability of LLMs has led to the development of explainable evaluation metrics~\cite{DBLP:conf/emnlp/Zhong0YMJLZJH22, DBLP:journals/corr/abs-2302-04166} to detect errors in LLM generations. \citet{DBLP:conf/emnlp/XuWPSFWL23} present InstructScore, an explainable metric based on LLaMA~\cite{DBLP:journals/corr/abs-2302-13971}, 
to obtain detailed error analysis for LLM-generated text. However, most of the current evaluation metrics require hard-to-obtain gold references.
\citet{jiang2023tigerscore} propose a reference-free evaluation metric, \textsc{TigerScore}, that can locate, categorize, and explain errors across various text generation tasks, including LFQA.
While LLM-based metrics can detect diverse errors, they are prone to hallucinations due to training data quality.
In this study, we collect expert-annotated data on fine-grained LFQA errors and train a feedback model for accurate error detection.

\vspace{-0.2cm}
\paragraph{Mitigating errors with human feedback.}
Reinforcement learning with human feedback~(RLHF)~\cite{DBLP:journals/corr/abs-1909-08593} incorporates human feedback to train reward models and align LLMs, reducing undesirable generations~\cite{DBLP:conf/nips/Ouyang0JAWMZASR22, DBLP:journals/corr/abs-2204-05862, DBLP:journals/corr/abs-2212-08073, DBLP:conf/nips/WuHSDSASOH23}.
A recent alignment technique, direct preference optimization~(DPO)~\cite{DBLP:conf/nips/RafailovSMMEF23}, bypasses the computationally expensive reward modeling step and has been used to fine-tune LMs for factuality~\cite{DBLP:journals/corr/abs-2311-08401}. 
Human feedback has also been used to fine-tune feedback models~\cite{DBLP:journals/corr/abs-2308-04592, xu2024llmrefine} to guide the refinement of LLM outputs~\cite{DBLP:conf/nips/MadaanTGHGW0DPY23, DBLP:conf/iclr/WelleckLWBSK023, DBLP:journals/corr/abs-2310-06271}, improving answer quality. 
However, these feedback models either lack fine-grained error feedback or depend on ground truth passages, which may not always be available in open-domain QA. In our study, we develop a reference-free feedback model to refine LFQA answers with detailed error feedback.

%% file: sections/4-data_collection.tex
Prior LFQA evaluations with non-expert~\cite{DBLP:journals/corr/abs-2112-09332} and expert~\cite{xu-etal-2023-critical} annotators collect preference judgments over model responses. However, overall preference is not indicative of fine-grained errors in LFQA. As a first step, we annotate span-level errors in long-form answers, with explanations from domain experts.

\subsection{Hiring Annotators}
\label{sec: hiring_annotators}

We recruit domain experts on Prolific's academic annotation platform for seven domains shown in \Cref{table:quantitative_results}. The expert selection is based on age (22-32), demographics (US and UK), education (undergraduate or graduate degree in the target domain), and native language (English). For each target domain, we first conduct a small pilot comprising ten samples, where given a question and two candidate answers, the experts evaluate the answers and mark all erroneous or problematic parts (phrase, sentence, or multiple sentences) based on our defined evaluation criteria (\cref{sec: task_setup}). After carefully evaluating the pilot results for relevance, clarity, and factuality, we choose three experts per domain and give them each a large-scale study containing 35-50 QA pairs. 
We collect expert judgments for 698 questions.

\begin{table}[!t]
\notsotiny
\centering
\begin{tabularx}{\linewidth}{@{}p{2.2cm}rrr@{}}
\toprule
\multirow{2}{2.5cm}{\textbf{Category} \\ \textbf{(\# samples)}} & \multicolumn{2}{c}{\textbf{Preference}} & \multirow{2}{2cm}{\textbf{Krippendorf's} \\ \raggedleft\textbf{$\alpha$}}\\
\cmidrule(lr){2-3}
& Human  & Model \\
\midrule
Physics ($94$) & $33 \%$ & $67 \%$ & $0.01$ \\
Chemistry ($96$) & $22 \%$ & $78 \%$ & $0.20$\\
Biology ($110$) & $25 \%$ & $75 \%$ & $0.36$ \\
Technology ($110$) & $16 \%$ & $84 \%$ & $0.53$ \\
Economics ($110$) & $14 \%$ & $86 \%$ & $0.31$ \\
History ($92$) & $9 \%$ & $91 \%$ & $0.52$ \\
Law ($86$) & $16 \%$ & $84 \%$ & $0.59$ \\

\midrule
\textbf{Average} & 19.29\% & 80.71\% & 0.36 \\

\bottomrule
\end{tabularx}
\caption{
Overview of HaluQuestQA and expert answer preferences, with experts' agreement on a smaller subset~(\textasciitilde$15\%$) calculated using Krippendorf's alpha~\cite{doi:10.1080/19312450709336664}~(\Cref{app:sec: exp_agreement}).
}
\label{table:quantitative_results}
\vspace{-0.2cm}
\end{table}

\subsection{Task Setup}
\label{sec: task_setup}

We evaluate two answers (human and model-generated) to the same questions. This setting enables us to identify errors made by humans and state-of-the-art LFQA systems. We chose GPT-4 (\texttt{gpt-4-0314}) as the LFQA model to evaluate since previous work~\cite{DBLP:journals/corr/abs-2309-08210} has shown it outperforming existing open-source LLMs (LLaMA and Alpaca~\cite{alpaca}) in reasoning and inferring from long context. Since GPT-4’s training data extends up to September 2021, it may have already seen the ELI5 dataset released by \citet{fan-etal-2019-eli5} during its pre-training. Thus, we scrape more recent questions with their highest-voted answers from the \emph{r/\text{explainlikeimfive}} subreddits posted between November 2022 and March 2023, following \citet{xu-etal-2023-critical}. We provide further details of the setup in \Cref{app: sec:data_analysis}.

To obtain the model-generated answers, we zero-shot prompt the GPT-4 model (\Cref{app: prompts: gpt_prompt}).
We host the annotation task on the INCEpTION platform~\cite{klie-etal-2018-inception} and evaluate the following:\footnote{We provide detailed annotation guidelines in \Cref{app: sec: annotation_guidelines}.}

\begin{enumerate}[left=0pt]
\setlength{\itemsep}{2pt}
 \item \textbf{Question misconception:} False assumptions made within the given question. 
 
 \item \textbf{Factuality:} 
 Accuracy and correctness of the answer as per verifiable facts.

 \item \textbf{Relevance:} 
 Specificity and meaningfulness of the answer.
 
\item \textbf{Completeness:} 
Answer comprehensiveness ensuring all question aspects are addressed.

 \item \textbf{References:} (Un)helpful examples, analogies, and external links in the answer. 

\end{enumerate}

\vspace{-0.2cm}

\noindent Annotators highlight errors in questions or answers based on evaluation criteria, provide free-form justifications, and indicate overall answer preferences.

\subsection{Data Curation}
\label{sec: data_curation}

To ensure high-quality annotations, we implement several checks before, during, and after the annotation process. 
Before annotation, we ensure the quality of scraped questions by removing duplicates, discarding questions under five words, and verifying grammar and fluency. 
For the corresponding answers, we select responses that receive the highest upvotes on Reddit and have a minimum length of 50 words. We also verify that the responses are not generated by AI-assisted tools (e.g., ChatGPT\footnote{\url{https://chat.openai.com}}) using a commercial AI detector (GPTZero\footnote{\url{https://app.gptzero.me/}}). We find that none of the responses are flagged as AI-generated, though we acknowledge the inherent limitations of AI detection tools. During annotation, annotators are encouraged to contact the authors anonymously through Prolific for clarification to reduce potential errors. After the study, we manually review error spans, justifications, references, and preference judgments, verifying their quality and ensuring no AI involvement.
Iterative feedback and bonus payments further incentivize high-quality work.
Examples are shown in \Cref{table:generations}.

\begin{table*}[!t]
\scriptsize
\centering
\begin{tabularx}{\textwidth}{p{1.2cm}p{2.3cm}XXX}
\toprule
\textbf{Aspect} & \textbf{Question} & \textbf{Answer A} & \textbf{Answer B}  & \textbf{Expert Justification}  \\
\midrule
\multirow{4}{*}{{\parbox{1.8cm}{Question\\Misconception}}}  & How do people in prison get so jacked with \hl{terrible nutrition and no supplements}?  
& \textit{\textcolor{green}{(human)}} [...] While prison food generally doesn't taste great, it does have calories and protein. [...] can buy on commissary (fish [...]
& \textit{(model)} [...] In prison, inmates often have a lot of free time, which they can use to work out and develop a consistent exercise routine. [...]
& This is an assumption, not every prison offers terrible food without nutritional value, also lots of "easy" food choices (lentils, eggs, [...] \\

\midrule

\multirow{4}{*}{{\parbox{1.4cm}{Factuality}}} &  What is difficult about developing nuclear weapons that makes some countries incapable of making them? & \textit{(human)}
The main problem is the nuclear fuel [...] \hl{Uranium is a fairly rare element on its own,} but to make a bomb you need lots of a very rare isotope of uranium (u-235) [...]   & \textit{\textcolor{green}{(model)}} 
Developing nuclear weapons is a highly complex and resource-intensive [...] To create a nuclear bomb, a nation must have a deep understanding of [...] &

The claim is false. According to the World Nuclear Association, Uranium is not a rare element and occurs in potentially recoverable concentrations in many types [...] \\

\midrule
\multirow{4}{*}{{\parbox{1.5cm}{Relevance}}} & Why is it taboo to talk about people's wages?  & \textit{(human)} 
Because companies used to be able to fire [...] discussing your wages is a federally [...] 
\hl{so by all means discuss discuss discuss!}~[...]  & \textit{\textcolor{green}{(model)}} 
Talking about wages is considered taboo because it can create tension and jealousy [...]. Employers often discourage discussions [...] & This does not contribute to the answer. \\

\midrule
\multirow{4}{*}{{\parbox{1.5cm}{Completeness}}} & Can anyone explain the differences between copyright and trademark? & \textit{(human)} A trademark protects a brand's symbol or logo. \hl{A copyright protects content.} So the ac/dc logo with the lightning bolt [...] 
& \textit{\textcolor{green}{(model)}}
[...] Copyright protects creative works like books, music, and art, [...]. Trademark, on the other hand, protects brand identifiers [...] & The answer fails to mention the broader scope of copyright protection, which includes creative works beyond just music and lyrics. \\

\midrule
\multirow{6}{*}{{\parbox{2.4cm}{References}}} & Why is 1000mbps internet only guaranteed at 900mbps and what's the equipment needed to get the 1000mbps advertised?  & \textit{(human)} [...] \hl{It's similar to asking why postal mail weighs more than just the paper you are sending... Everything that tells it how to get to the other side encapsulates (contains) it, and makes up that extra weight.} & \textit{\textcolor{green}{(model)}} Internet speeds are often advertised as "up to" a certain speed because various factors can affect the actual performance. For instance, network congestion... To get the full 1000mbps advertised, [...]  & The example is terrible and does not give any idea of what the writer is talking about.\\

\bottomrule
\end{tabularx}
\caption{Examples of expert annotated \hl{errors} in long-form answers based on the defined evaluation criteria. The answers in \textcolor{green}{\textit{green}} are expert-preferred answers. Additional examples are provided in \Cref{table:additional_generations} in the Appendix.}
\label{table:generations}
\end{table*}

\subsection{Quantitative Analysis}
\label{sec: quant_analysis}

As shown in \Cref{table:quantitative_results}, experts display a high preference~(80.7\%) for GPT-4 answers compared to human answers. 
We hypothesize that humans prefer fluent answers, and LLMs are known to optimize for fluency~\cite{DBLP:journals/corr/abs-2303-13648, DBLP:journals/corr/abs-2303-14342}.
Moreover, the preference of our annotators is corroborated by similar findings in summarization~\cite{DBLP:conf/acl/LiuFLZNHHJWXR23} and LFQA~\cite{xu-etal-2023-critical}, who show that GPT-3 answers score higher than human answers.

Comparing different domains, we observe that experts strongly prefer GPT-4 answers in history, law, technology, and economics (>80\%). However, in science domains like physics, biology, and chemistry, model preference drops to 60-80\%.
GPT-4's strong performance in history and law can be attributed to its ability to learn facts that remain consistent over time, during its training process. 
However, it struggles with college-level scientific questions requiring advanced reasoning~\cite{DBLP:conf/aaai/SunHZMSC0024, DBLP:conf/icml/WangHL0ZSLZS024} -- and our dataset includes complex, real-world scientific problems that surpass college-level difficulty, likely contributing to its lower performance in scientific domains.

\subsection{Fine-grained Answer Scoring} 
We score human and model answers on our defined evaluation criteria to understand how experts' answer preferences diverge across different domains.
For the \textit{question misconception} aspect, the score $\mathcal{S}=1$ when the question has no misconceptions; otherwise, $\mathcal{S}=0$. For aspects of \textit{factuality}, \textit{relevance}, and \textit{completeness}, $\mathcal{S} = 1 - \left( \frac{{\text{{\# Error sentences}}}}{{\text{{Total \# of sentences}}}} \right)$, while the score for \textit{reference} is calculated as $\mathcal{S} = 1 - \left( \frac{{\text{{\# Error references}}}}{{\text{{Total \# of references}}}} \right)$.

\begin{figure}[!t]
\centering
\includegraphics[scale=0.15]{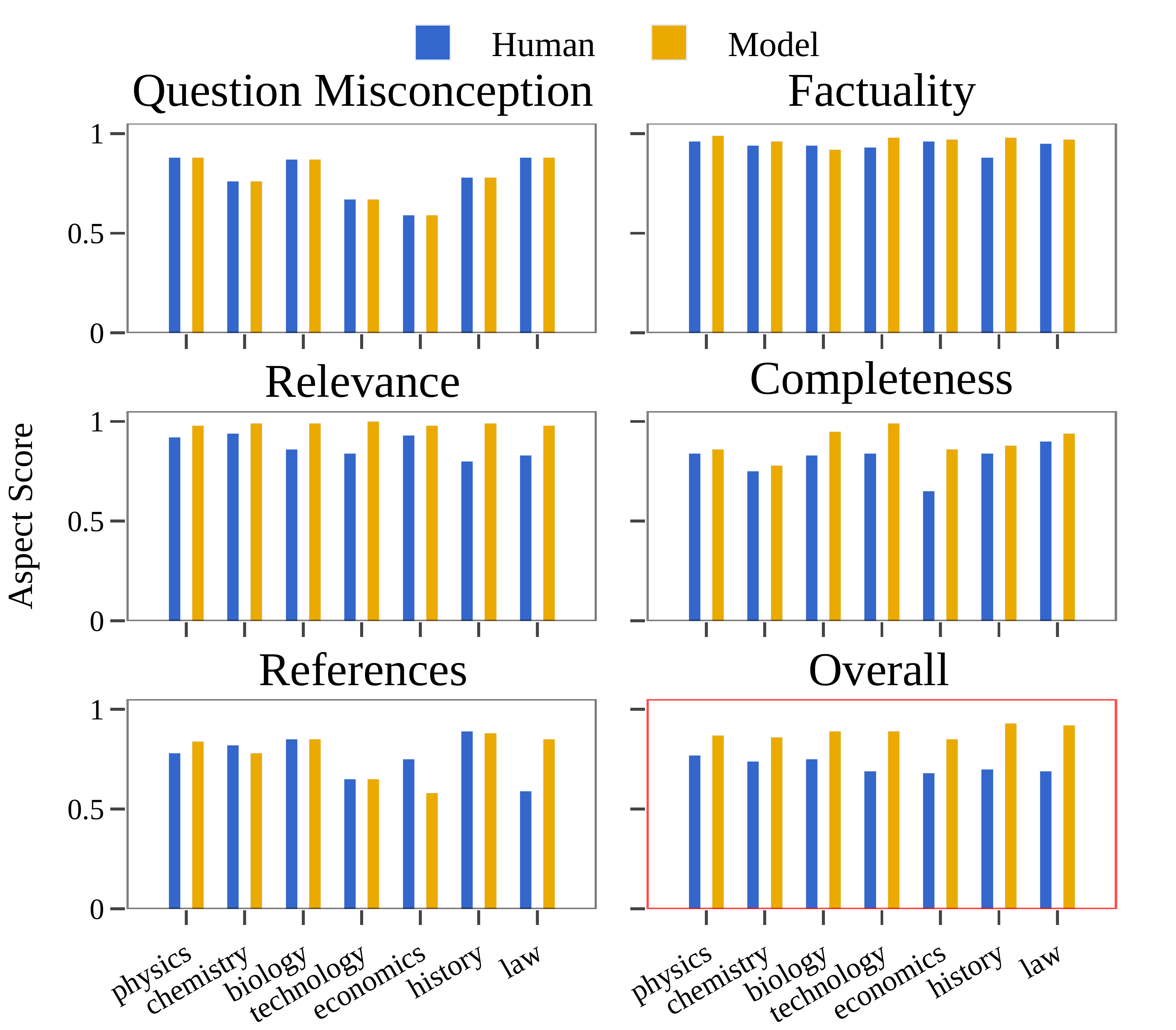}
\caption{Comparison of fine-grained scores of the human-written and model-generated answers for different evaluation criteria. The last figure (with \textcolor{red}{red} boundary) shows the averaged and normalized overall scores. A higher score represents fewer errors in the answers.}
\label{fig:aspect_scores}
\vspace{-0.2cm}
\end{figure}

\noindent For calculating the overall answer scores, we leave out the question misconception scores because this aspect pertains to the question. 
We sum the other aspect scores and include the overall answer preference scores ($\mathcal{S}=1$ if preferred) to get the final score.
Finally, we normalize this score between 0 and 1.
In \Cref{fig:aspect_scores}, we report the fine-grained aspect scores for human and model answers across different domains and discuss our findings below.

\paragraph{1) Questions from technology and economics are biased.}
Ambiguous and misinformed questions can lead to undesirable answers~\cite{DBLP:conf/emnlp/ColeZGEDE23, DBLP:conf/emnlp/KimKJPK23}. Therefore, fair answer scoring requires prior estimation of question quality.
For this, we utilize the question misconception aspect and find that questions from all evaluated domains consist of misconceptions arising from the user's bias or misinformation. This is especially prominent in technology and economics, where \textasciitilde$40\%$ of the questions are misinformed -- users have low domain knowledge to ask the right questions.
Thus, we encourage future research to assess the capability of LLMs to rebut misconceived questions.

\paragraph{2) Answers lack comprehensiveness and provide unhelpful references.} We observe that human-written and model-generated answers score high on \textit{factuality} and \textit{relevance}, meaning most of the information provided is verifiable, trustworthy and relevant to the question. However, the answers score low on \textit{completeness} and \textit{references} aspects, lacking important information and providing web references and examples that are not helpful~\cite{DBLP:conf/emnlp/LiuZL23}, according to expert judgments. Specifically, GPT-4 hallucinates and provides incorrect or fabricated web links, while human answers digress from the topic and include irrelevant information.

Overall, GPT-4 answers score better than the human answers in all evaluated domains. 
While this is due to its better performance over humans in the considered aspects, the persuasive nature of the model responses~\cite{DBLP:journals/corr/abs-2403-14380} also plays a crucial role in their higher preference.

%% file: sections/5-hallucination.tex
In \Cref{sec: quant_analysis}, we have shown that the LFQA answers lack completeness and omit useful information. Therefore, we train a feedback model to identify erroneous answer spans with \textit{incomplete information} and provide free-form error justifications. 
Our approach \textsc{Error-informed refinement}, uses this feedback to refine the answers and improve their overall quality without human intervention. 

\begin{figure*}[!t]
\centering
\begin{tikzpicture}[node distance=1cm, auto,
  block/.style={rectangle, line width=0.5mm, fill=white, rounded corners, minimum height=3cm, font=\scriptsize, align=left},
  redblock/.style={block, draw=red, text width=3.5cm},
  yellowblock/.style={block, draw=orange, text width=3.5cm},
  greenblock/.style={block, draw=green, text width=3.5cm},
  arrow/.style={->, thick}
]

\node[redblock] (initial) {%
  \textbf{Q:} Why are railroads full of rocks?\\[0.3em]
  \textbf{A:} Ballast is a comparably cheap way of leveling ground that can support a lot of weight very cheaply. Rocks don't wash or blow away [...]. Compared to say a concrete foundation, they won’t crack or require formwork [...]
};

  \node (feedback) [right=0.3cm of initial, font=\small] {%
    \begin{tikzpicture}
      \node at (-1,-0.6) {\LARGE\texttwemoji{robot}}; 
      \node[align=center] at (-1.4,-1.5) {Feedback\\Model\\$\mathcal{E}$};
    \end{tikzpicture}
  };

\node[yellowblock] (error) [right=0.3cm of feedback] {%
    \textbf{Consistency score ($S_{RC}$)}: 0.8\\[0.3em]
    \textbf{Prediction (Sentence-level)}:\\[0.1em]
1. \hl{[Incomplete] Reasons: This is correct; ballast is cheap and effective [...]. However, I think it will be useful to mention that ballast also helps with drainage and prevents erosion.} \\
2. [Complete] \\
3. [Complete]
};

  \node (refine) [right=0.3cm of error, font=\small] {%
    \begin{tikzpicture}
      \node at (-1.2,-0.6) {\LARGE\texttwemoji{robot}}; 
      \node[align=center] at (-1.4,-1.5) {Refine\\Model\\$\mathcal{R}$};
    \end{tikzpicture}
  };

\node[greenblock] (final) [right=0.3cm of refine] {%
  \textbf{Refined Answer}\\[0.5em]
  Railroads are full of rocks because ballast provides an economical and efficient solution for leveling ground and \textbf{preventing erosion}. Additional benefits include thermal cycling, self-leveling, and ease of repair. [...] Here are some reasons [...] 1. Cost effective [...] 2. \textbf{Drainage} [...] 3. Thermal [...]
};

\draw[arrow] (initial) -- (feedback);
\draw[arrow] (feedback) -- (error);
\draw[arrow] (error) -- (refine);
\draw[arrow] (refine) -- (final);
\end{tikzpicture}
\caption{A pictorial view of our Error-Informed Refinement approach. The \textsc{feedback} model takes a question-answer pair as input and outputs sentence-level error with justifications and a consistency score. The \textsc{refine} model uses this feedback to improve the original answer.  Additional refined examples are in \Cref{table:eir_examples} (Appendix).}
\label{fig: eir}
\vspace{-0.2cm}
\end{figure*}

\subsection{Error Feedback Model}
\label{sec: error_feedback}

Given an input question and an LFQA response, we fine-tune the LLaMA2-13B model~\cite{DBLP:journals/corr/abs-2307-09288} to generate a label \textit{[Complete]} or \textit{[Incomplete]} for every sentence $1...n$ in the response and provide associated reasons for the incomplete sentences~(see \Cref{fig: eir}).

\vspace{-1mm}
\paragraph{Dataset \& Training.}

Training the feedback model requires high-quality error annotations with justifications. To support this, we extract QA pairs with annotated completeness errors from our dataset, which includes both sentence- and phrase-level annotations.
Since \textasciitilde$65\%$ of annotated completeness errors occur at the sentence level~(see \Cref{table:span_level_stats}), we adopt sentence-level granularity as both representative and practical for training and evaluating our feedback model.
To facilitate this, we convert phrase-level errors to sentence-level annotations by assigning the phrase's error label and justification to its containing sentence. 
Furthermore, in cases where annotators mark the entire answer as incomplete~(\Cref{app:subsec: task_instructions}), we assign the same error label and justification to each sentence in the answer. We provide illustrative examples of sentence-level annotations in \Cref{table: feedback_data_train} in the Appendix.

Finally, after preprocessing, we segment each human- or model-generated answer into sentences and label each sentence as \textit{[Complete]} or \textit{[Incomplete]}, with the corresponding expert-provided justification.
The final dataset consists of $509$ samples with a $90/10$ train/test split. 
Fine-tuning on this dataset is essential for accurate error feedback generation, as general-purpose LLMs used in a zero-shot setting are not well-suited to detecting completeness errors~(\cref{subsec: detect_errors}).
We provide the training details in \Cref{app: prompts: feedback}.

\vspace{-1mm}
\paragraph{Inference.}

The trained feedback model hallucinates web references in about 20\% of test samples. This likely occurs because the training data includes web references in expert error justifications, which the model struggles to replicate coherently.
To combat this, we opt for a sampling-based approach~\cite{DBLP:journals/corr/abs-2403-00696} to provide more consistent feedback. The intuition is that trustworthy details and references should appear in many other generated samples. Hence, during the decoding step, we
sample 20 responses from the feedback model and check their consistency in two stages: \underline{(1)} \textsc{Tag consistency}: This pertains to the consistency of span-level tag predictions, \textit{complete} or \textit{incomplete}, for each sampled response. The tag consistency score is calculated by counting the number of other sampled responses that match the tag sequence of each sampled output and averaging over the total number of samples. Formally, if the sampled tag predictions $p_1, ..., p_n$ consist of tag sequences $t_1, ..., t_n$ where $t_i$ is a list of tag predictions for every span, the score for sample $i$ is

\[
\boxed{
\small
\begin{aligned}
    \mathcal{S_{TC}} = \frac{1}{n} \sum_{s=1}^n 1_{t_i=t_s}
\end{aligned}
}
\]

\noindent where $1_{t_i=t_s}$ is 1 if the tag sequence $t_i$ is the same as tag sequence $t_s$ and 0 if not.
\textbf{The samples with the highest score are selected for the next stage.}
\underline{(2)} \textsc{Reason consistency}: We assess the consistency of justifications given for the incomplete spans from the remaining samples. Specifically, we count the number of other sampled justifications from the LLM that matched each token of each sampled output and score each justification by the average count per token. Formally, if the sampled justifications $j_1, ..., j_n$ consist of words $w_i^k, k=1 ... m_i$, the score of sample $i$ is

\[
\boxed{
\small
\begin{aligned}
     \mathcal{S_{RC}} = \frac{1}{m_i} \sum_{k=1}^{m_i} \sum_{s=1}^n 1_{w_i^k \in j_s}
\end{aligned}
}
\]

\noindent where $1_{w_i^k \in j_s}$ is 1 if token $w_i^k$ is in the justification $j_s$ and 0 if not. 
Finally, we select the highest scoring output as feedback for the refinement model. After sampling,
\textbf{reference hallucinations reduce by 50\% (from 20\% to \textasciitilde10\% of the test set).}

\subsection{Error-Informed Refinement~(EIR)}
\label{sec: error-informed refinement}
Our approach is shown in \Cref{fig: eir} and consists of two main components: an error feedback model~(\cref{sec: error_feedback}), and a refinement model. Given an input prompt $x_i$ and a corresponding human-written or model-generated response $y_i$, the feedback model $\mathcal{E}$ generates a targeted feedback $f_i$ that represents the quality of $y_i$ in free-form natural language. Finally, the refinement model $\mathcal{R}$ uses $x_i$, $y_i$, and $f_i$ to generate a refined and improved output $\hat{y}_i$. The following sections describe our approach.

\paragraph{Refinement Model \& Baselines.}
Our experiments use the LLaMA2-13B chat LLM and its DPO optimized version~(see \Cref{app: sec: dpo}) 
as the refinement model. In each case, the model is 0-shot prompted with the fine-grained error feedback received from the error detection model. We also experiment with two strong baseline feedback models, \underline{(1)} \textbf{\textsc{Improve}}: The refinement model is 0-shot prompted to improve the answer without any error feedback provided. \underline{(2)} \textbf{\textsc{Generic}}: The refinement model is 0-shot prompted to improve the answer with a generic error feedback that asks the model to provide a more complete and accurate answer. 
We list the prompts used in \Cref{app: prompts: refinement}.

\paragraph{Datasets \& Evaluation Metrics.}
\label{sec: data_and_eval}

We test our error-informed refinement approach on three datasets: \textsc{HQ$^2$A} with span-level error annotations for answer completeness, ASQA~\cite{stelmakh-etal-2022-asqa}, and ELI5~\cite{fan-etal-2019-eli5}. The ASQA dataset consists of 6K ambiguous factoid questions with long-form answers synthesized from multiple sources to resolve the ambiguities. ELI5 consists of 270K long-form answers covering general topics from the Reddit forum "Explain Like I'm Five".

We evaluate the refined answers using TIGERScore, a trained reference-free metric that identifies errors in LLM-generated text and assigns an \textit{error score} based on error severity.
Specifically, we use the LLaMA-7B trained version of TIGERScore, which highly correlates with humans for error detection in LFQA tasks~\cite{jiang2023tigerscore} while being cost-effective.
We also measure how well our refinement approach corrects errors identified by TIGERScore using precision, recall, and F1 score metrics. 
Finally, we conduct a human evaluation to assess the comprehensiveness and preference of refined answers compared to gold answers.

%% file: sections/6-results.tex
We explore several research questions: 
\underline{(1)} Can our learned feedback model detect errors in LFQA systems and help in downstream answer refinement task?
\underline{(2)} Does fine-grained feedback produce better quality LFQA answers than coarse-grained feedback?
\underline{(3)} Does fine-grained feedback help mitigate errors and improve the comprehensiveness of LFQA answers?
\underline{(4)} Are comprehensive answers from our approach preferred by humans?

\subsection{Detecting Errors via Feedback Model}
\label{subsec: detect_errors}

\begin{table*}[t!]
\notsotiny
\centering
\begin{tabularx}{\textwidth}{@{}llXXXXX@{}}
\toprule
\multirow{2}{*}{\centering\textbf{Approach}} & \multirow{2}{*}{\centering\textbf{Model}} & \multicolumn{3}{c}{\textbf{Accuracy (\%)}} & \multirow{2}{*}{\shortstack[l]{\textbf{Weighted} \\ \textbf{Accuracy (\%)~($\uparrow$)}}} & \multirow{2}{*}{\shortstack[l]{\textbf{Consistency} \\ \textbf{Score~($S_{RC}$)~($\uparrow$)}}} \\
\cmidrule(lr){3-5} 
 & & \textbf{Exact ($\uparrow$)} & \textbf{Adjacent ($\downarrow$)} & \textbf{Different ($\downarrow$)} \\
\midrule

Zero-shot & LLaMA2-13B & $23.53 \pm 1.60$ & $\textbf{7.84} \pm 0.00$ & $68.63 \pm 1.60$  & $34.31 \pm 1.44$ & $0.52 \pm 0.02$ \\
Zero-shot & GPT-3.5-Turbo & $25.49$ & $11.76$ &   $62.75$  & $37.65$ &  $\textbf{0.99}$   \\
Fine-tuning w/ HQ$^2$A & LLaMA2-13B & $\textbf{37.25} \pm 0.00$ & $24.18 \pm 0.92$ &    $\textbf{38.56} \pm 0.93$ & $\textbf{53.20} \pm 0.37$ & $0.80 \pm 0.01$  \\
\bottomrule
\end{tabularx}
\caption{
 Accuracy and Consistency Score ($S_{RC}$) of zero-shot and fine-tuned models in detecting sentence-level errors on \textsc{HQ$^2$A}, averaged over three runs with standard deviations~(except for GPT-3.5). Best scores are in \textbf{bold}.
}
\label{table:error_detection}
\vspace{-0.2cm}
\end{table*}

To measure the error detection accuracy of our feedback model, we propose an evaluation across three fine-grained categories:
\underline{(1)} \textbf{\textsc{Exact}}: Erroneous sentences identified by the model exactly match the human-annotated erroneous sentences. This category represents the most stringent evaluation of model performance.
\underline{(2)} \textbf{\textsc{Adjacent}}: Erroneous sentences identified by the model are adjacent to, or closely related to, human-annotated erroneous sentences. Here, "adjacent" refers to a sentence preceding or following the human-annotated error sentence. These near-misses may still aid in understanding or resolving the error due to the contextual relation between the preceding/following sentence and the actual error sentence.
\underline{(3)} \textbf{\textsc{Different}}: Erroneous sentences identified by the model do \textit{not} match, precede or follow human-annotated error sentences, capturing instances where the model detects completely unrelated error spans.

To capture the overall error detection performance across the defined evaluation categories, we introduce a \textbf{weighted accuracy} metric:

\[
\boxed{
\scriptsize
\text{Accuracy}_{\text{wt}} = 
\begin{aligned}
& w_{\text{Exact}} \cdot \frac{\text{\# Exact matches}}{\text{Total errors}}
+ w_{\text{Adj}} \cdot \frac{\text{\# Adjacent matches}}{\text{Total errors}} \\
& + w_{\text{Diff}} \cdot \frac{\text{\# Different matches}}{\text{Total errors}}.
\end{aligned}
}
\]

\vspace{-2mm}
\noindent where $w_{\text{Exact}}$, $w_{\text{Adj}}$, and $w_{\text{Diff}}$ represent the weights assigned to each category according to its relative importance. We assign $w_{\text{Exact}}$ = 1.0 to reward the model's capability of correctly detecting errors and $w_{\text{Adj}}$ = 0.5 to quantify the importance of near-misses which may still provide insight on the actual errors. Consequently, $w_{\text{Diff}}$ = 0.1 to penalize the model for its incorrect error detection.

In \Cref{table:error_detection}, we show the sentence-level error detection accuracy of the zero-shot LLaMA2-13B and GPT-3.5-Turbo and our fine-tuned feedback models compared to the strong human baseline. 
Our fine-tuned feedback model improves the detection of correct error spans~(\emph{exact}) by \textasciitilde$14\%$ and \textasciitilde$12\%$ and reduces the detection of incorrect error spans~(\emph{different}) by \textasciitilde$30\%$ and \textasciitilde$24\%$ compared to the zero-shot feedback models LLaMA2-13B and GPT-3.5-Turbo, respectively.
Specifically, our feedback model outperforms GPT-3.5-Turbo by \textasciitilde$16\%$ on our weighted accuracy metric while maintaining a high consistency score of $0.80$. 
This shows that the model effectively learns to identify completeness errors, even when fine-tuned on \textit{limited but high-quality} \textsc{HQ$^2$A} samples, aligning with recent findings~\cite{DBLP:conf/nips/ZhouLX0SMMEYYZG23, DBLP:conf/icml/XiaMGA024} on fine-tuning with small but carefully curated datasets. In \Cref{app: prompts: feedback}, we discuss systematic patterns of errors learned by the feedback model.

We further evaluate our error feedback model by comparing the gap in the downstream LFQA refinement task when we use human-annotated error feedback. 
This evaluation measures the effectiveness of our feedback model in guiding the refinement of long-form answers and reducing errors. 
In \Cref{table:feedback_refine}, we present the refinement performance with our feedback model compared to the expert human feedback on \textsc{HQ$^2$A}. 
Our feedback model reduces error samples by $2\%$ and improves the F1 score by $4\%$ over expert human feedback, validating its effectiveness in refining LFQA answers.

\begin{table*}[htbp]
\notsotiny
\centering
\renewcommand{\arraystretch}{1.2} 
\begin{tabularx}{\textwidth}{@{}p{1cm}lll|XXX@{}}
\toprule
\multirow{3}{*}{\textbf{Dataset}} & \multirow{3}{*}{\textbf{Approach}} & \multicolumn{2}{c}{\textbf{TIGERScore}} & \multicolumn{3}{c}{\textbf{Error Correction}} \\
\cmidrule(lr){3-4} \cmidrule(lr){5-7} 
& & \multicolumn{1}{l}{\textbf{\% Error samples ($\downarrow$)}} & \multicolumn{1}{l}{\textbf{Error score ($\downarrow$)}} & \multicolumn{1}{l}{\textbf{Precision ($\uparrow$)}} & \multicolumn{1}{l}{\textbf{Recall ($\uparrow$)}} & \multicolumn{1}{l}{\textbf{F1 ($\uparrow$)}} \\

\midrule

\multirow{2}{1.5cm}{\textsc{HQ$^2$A}}  & Human feedback & $2.61\pm0.92$  & $0.09\pm0.01$ &  $0.86\pm0.04$ &  \greenbox{$\textbf{1.00}\pm0.00$} & $0.94\pm0.02$  \\
\cdashline{2-7}
& Baseline &  $19.61$  & $0.63$ & - & - & -  \\
& Zero-shot &  $15.69\pm0.00$  & $0.34\pm0.00$ & $0.56\pm0.00$ & $0.90\pm0.00$ & $0.69\pm0.00$  \\
 &  Improve & \orangebox{$1.31\pm0.92$}  & $0.05\pm0.04$ & \greenbox{$\textbf{1.00}\pm0.00$} & $0.93\pm0.05$ & \orangebox{$0.97\pm0.02$} \\
 &  Generic & \orangebox{$1.31\pm0.92$}  & \orangebox{$0.05\pm0.03$} & \orangebox{$0.97\pm0.04$}  & \orangebox{$0.97\pm0.05$}  & \orangebox{$0.97\pm0.02$} \\

 \cdashline{2-7}
 & \textsc{EIR} (\textit{Ours}) & \greenbox{$\textbf{0.65}\pm0.92$}  & \greenbox{$\textbf{0.03}\pm0.04$} & \orangebox{$0.97\pm0.04$} &  \greenbox{$\textbf{1.00}\pm0.00$} & \greenbox{$\textbf{0.98}\pm0.02$}  \\
   
\midrule

\multirow{2}{1.5cm}{\textsc{ASQA}}  &  Baseline & $34.81$ & $1.20$ & - & - & - \\
& Zero-shot &  $35.02\pm0.00$  & $1.08\pm0.00$ & $0.50\pm0.00$ & $0.62\pm0.00$ & $0.55\pm0.00$  \\
 &  Improve & $20.85\pm1.00$  & $0.68\pm0.03$ & $0.70\pm0.02$ &  $0.71\pm0.01$ & $0.70\pm0.01$ \\
 &  Generic &  \orangebox{$18.67\pm0.52$} & \orangebox{$0.61\pm0.01$} & \orangebox{$0.72\pm0.01$} &  \orangebox{$0.75\pm0.01$} & \orangebox{$0.74\pm0.00$} \\

   \cdashline{2-7}
 & \textsc{EIR} (\textit{Ours}) &  \greenbox{$\textbf{16.63}\pm0.41$}  & \greenbox{$\textbf{0.51}\pm0.02$} & \greenbox{$\textbf{0.73}\pm0.00$} & \greenbox{$\textbf{0.82}\pm0.02$} & \greenbox{$\textbf{0.77}\pm0.01$} \\

\midrule

\multirow{2}{1.5cm}{\textsc{ELI5}}  &  Baseline & $22.93$   & $0.82$ & - &  - & -  \\
& Zero-shot &  $9.61\pm0.00$  & $0.27\pm0.00$ & $0.74\pm0.00$ & $0.89\pm0.00$ & $0.81\pm0.00$  \\
 &  Improve & $10.05\pm0.18$  & $0.36\pm0.02$  & $0.75\pm0.00$  & $0.86\pm0.00$ & $0.80\pm0.00$  \\
 &  Generic & \orangebox{$6.06\pm0.23$}  & \orangebox{$0.22\pm0.01$}  & \orangebox{$0.84\pm0.01$} & \orangebox{$0.91\pm0.00$} & \orangebox{$0.87\pm0.00$} \\

 \cdashline{2-7}
 & \textsc{EIR} (\textit{Ours}) &  \greenbox{$\textbf{3.81}\pm0.30$} &  \greenbox{$\textbf{0.13}\pm0.01$} & \greenbox{$\textbf{0.88}\pm0.01$}   &  \greenbox{$\textbf{0.96}\pm0.01$}  & \greenbox{$\textbf{0.92}\pm0.01$}  \\
 

\bottomrule
\end{tabularx}
\caption{
Results on the quality of original answers from the datasets (\textsc{Baseline}); answers from zero-shot prompting LLaMA2-13B-chat (\textsc{Zero-shot}); answers refined with coarse-grained feedback (\textsc{Improve} and \textsc{Generic}), fine-grained feedback (\textsc{EIR}) and human feedback on \textsc{HQ$^2$A}. 
Reported results are averages over three iterations with standard deviations. 
Best results are in \textbf{bold green}, and the second-best results are in orange. We report results with LLaMA3-8B-Instruct~\cite{DBLP:journals/corr/abs-2407-21783} and Mistral-7B-Instruct-v0.3~\cite{DBLP:journals/corr/abs-2310-06825} models in \Cref{app: additional_refinement_models}.
}
\label{table:feedback_refine}
\vspace{-0.2cm}
\end{table*}

\subsection{Fine- vs. Coarse-Grained Feedback}
\label{sec: results_feedback}

\Cref{table:feedback_refine} presents the quality of \textsc{Baseline} answers (original dataset instances) refined using coarse- and fine-grained feedback. We also evaluate answers generated through zero-shot prompting LLaMA2-13B-chat for comparison.

Our results show that inadequate feedback can deteriorate generation quality. While directly prompting the refinement model to generate answers (\textsc{Zero-shot}) or improve answers without detailed feedback (\textsc{Improve}) performs better than the baseline, using more targeted feedback, such as asking the model to complete the answer (\textsc{Generic}), consistently leads to higher-quality LFQA answers.
In contrast, fine-grained feedback from our error detection model~(\textsc{EIR}) outperforms coarse-grained feedback and fine-grained human feedback (on \textsc{HQ$^2$A}), reducing error samples and error scores by \textasciitilde$3\%$ and  \textasciitilde$\Delta38\%$, respectively, and improving F1 scores by \textasciitilde$5\%$, on average.

We also investigated the impact of aligning the refinement model with human preferences from \textsc{HQ$^2$A} with DPO. Despite promising initial results in reducing LFQA errors (\Cref{app: align_lms}), the resulting refinement model ultimately did not outperform the vanilla refinement model (\Cref{app: sec: eir_dpo}).

\subsection{Human Evaluation}
We conduct a human evaluation with three annotators to test the completeness and overall quality of the answers generated using our refinement approach. 
For 50 questions each from the HQ$^2$A, ASQA, and ELI5 datasets, we present annotators with a pair of answers—one from the dataset (baseline) and one refined by our method.

To evaluate completeness, we adopt a comparative \textit{comprehensiveness} metric: annotators judge which answer more fully addresses all parts of the question, based on our defined criteria for identifying completeness errors~(see~\Cref{app:subsec: task_instructions}). To assess the overall answer quality, annotators consider broader factors, such as the factual precision and relevance~(\Cref{app:subsec: task_instructions}), when selecting their preferred answer.

\Cref{table:human_eval_results} shows the results of our human evaluation of the \textsc{baseline} and \textsc{refined} answers. We observe that refined answers are considered more comprehensive in \textasciitilde60\% of cases and preferred overall in \textasciitilde84\% of comparisons on average across all evaluated datasets, demonstrating improved completeness and quality over the baseline answers.

\begin{table}[!t]
\notsotiny
\centering
\begin{tabularx}{0.48\textwidth}{@{}>{\raggedright\arraybackslash}p{0.12\linewidth}p{0.13\linewidth}>{\centering\arraybackslash}p{0.36\linewidth}>{\centering\arraybackslash}p{0.22\linewidth}@{}}
\toprule
\textbf{Dataset} & \textbf{Pref.} & \textbf{Comprehensiveness}\textsuperscript{($\uparrow$)} & \textbf{Overall}\textsuperscript{($\uparrow$)} \\
\midrule
\multirow{2}{1.5cm}{\textsc{HQ$^2$A}} & Baseline & $0.00 \%$ & $7.84 \%$ \\
 & Refined & $\textbf{100} \%$ & $\textbf{92.16} \%$   \\
& Tie & $0.00 \%$ & -  \\
\midrule
\multirow{2}{1.5cm}{\textsc{ASQA}} & Baseline & $0.00 \%$ & $40.00 \%$ \\
 & Refined & $18.00 \%$ & $\textbf{60.00} \%$  \\
 & Tie & $\textbf{82.00} \%$ & -  \\
\midrule
\multirow{2}{1.5cm}{\textsc{ELI5}} & Baseline & $0.00 \%$ & $0.00 \%$ \\
 & Refined & $\textbf{62.00} \%$ & $\textbf{100} \%$ \\
  & Tie & $38.00 \%$ & -  \\
\bottomrule
\end{tabularx}
\caption{
Human evaluation results on the comprehensiveness and preference of answers refined with \textsc{EIR} over the original answers from the datasets~(\textsc{Baseline}). 
Details on the human agreement are in \Cref{app: human_eval}.
}
\label{table:human_eval_results}
\vspace{-0.3cm}
\end{table}

%% file: sections/7-conclusion.tex
We introduce \textsc{HaluQuestQA}, a dataset of expert judgments on fine-grained errors in LFQA. 
Using our dataset, we analyze the pitfalls of human and model long-form answers, identifying issues with comprehensiveness and unhelpful references. To address these, we propose \textsc{Error-informed refinement}, an approach that uses signals from our learned feedback model to refine LLM responses. Our feedback model outperforms baseline feedback models and expert human feedback in guiding answer refinement and reducing errors. A human evaluation confirms the effectiveness of our approach, with participants finding our refined answers more comprehensive and preferable to baseline outputs.

%% file: sections/8-limitations.tex
Despite providing an in-depth analysis on errors in human and model generated responses, our work only focusses on the LFQA task. Thus, we encourage future work to apply our findings to different tasks such as summarization, translation, etc.
We study a diverse but limited scope of long-form answers drawn from online community platforms. More diverse questions from different domains such as education or commercial may have different issues and might be to be evaluated in a different way.

Our trained error detection model shows high correlation with human annotations but relies on a high consistency of model outputs. The model may hallucinate if the consistency score is low ($<0.80$). Training larger models with more high quality data might be an interesting future work to get better results.
Lastly, in our refinement approach, we have experimented with the instruction-tuned variants of the LLaMA2, LLaMA3, and Mistral models. 
Models with better or worse instruction following capabilities may give different results, and improving the refinement process can be a great future direction to mitigate errors.

%% file: sections/9-ethics.tex
The expert annotation data collection protocol has been determined to be exempt from review by an IRB board. All the collected data will be publicly available under the CC BY-SA 4.0 license.
We hire annotators on the academic annotation platform Prolific and gather no sensitive user information except demographics and annotator performance data.
We examined the collected data and ascertained that it contains no toxic or harmful content. 

%% file: sections/acknowledgements.tex
This research work is funded by the German Research Foundation (DFG) as part of the UKP-SQuARE project (grant GU 798/29-1) and by the German Federal Ministry of Education and Research and the Hessian Ministry of Higher Education, Research, Science and the Arts within their joint support of the National Research Center for Applied Cybersecurity ATHENE. Yixiao Song and Mohit Iyyer are supported by the award IIS-2312949 from the National Science Foundation (NSF).

We thank Sukannya Purkayastha and Haritz Puerto for their insightful feedback on the paper and Manika Arvind Arora for the valuable feedback on the annotation setup.
Lastly, we are grateful to our dedicated annotators who helped create the HaluQuestQA dataset.

%% file: sections/10-appendix.tex
\section{Data Collection and Analysis}
\label{app: sec:data_analysis}
This section presents additional insights on our HaluQuestQA~(HQ$^2$A) dataset.

\subsection{Domain Classification}
The questions on the ELI5 are classified into domains via the FLAIR label~(tag containing post information), which lets us perform domain-specific analysis. 
For unclassified categories~(like History and Law), we cluster the OTHER category questions~(not in pre-defined ELI5 domains), using K-means clustering~\cite{10.1109/TPAMI.1984.4767478} and identify the domain-specific questions. For each domain, we sample between 100-200 questions with their highest-voted answers.

\subsection{Answer Length Distribution}
\begin{figure}[!h]
\includegraphics[width=0.45\textwidth]{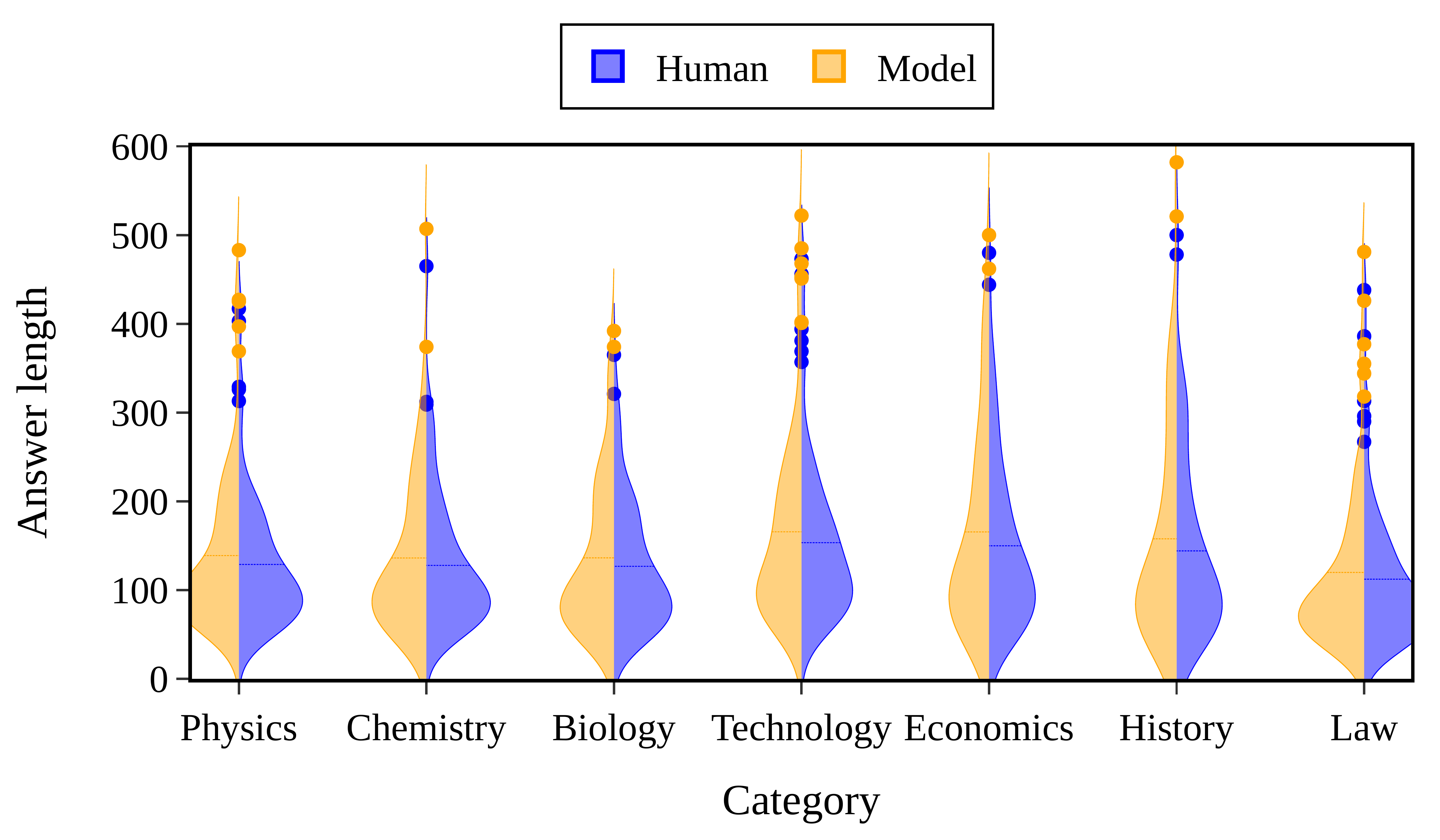}
\caption{Answer length distribution of human-written and model-generated answers (H/M) in our expert-annotated dataset.}
\label{fig:ans_length}
\end{figure}

\Cref{fig:ans_length} compares the length distribution of human-written and model-generated answers.
We observe that the length of human and model answers is comparable, resulting in a fair evaluation.
Across all domains, the length of collected answers ranges between 50-500 words with an average length of 100 words. 

\subsection{Overall Answer Preference}
\begin{figure*}[!t]
\centering
\includegraphics[width=\textwidth]{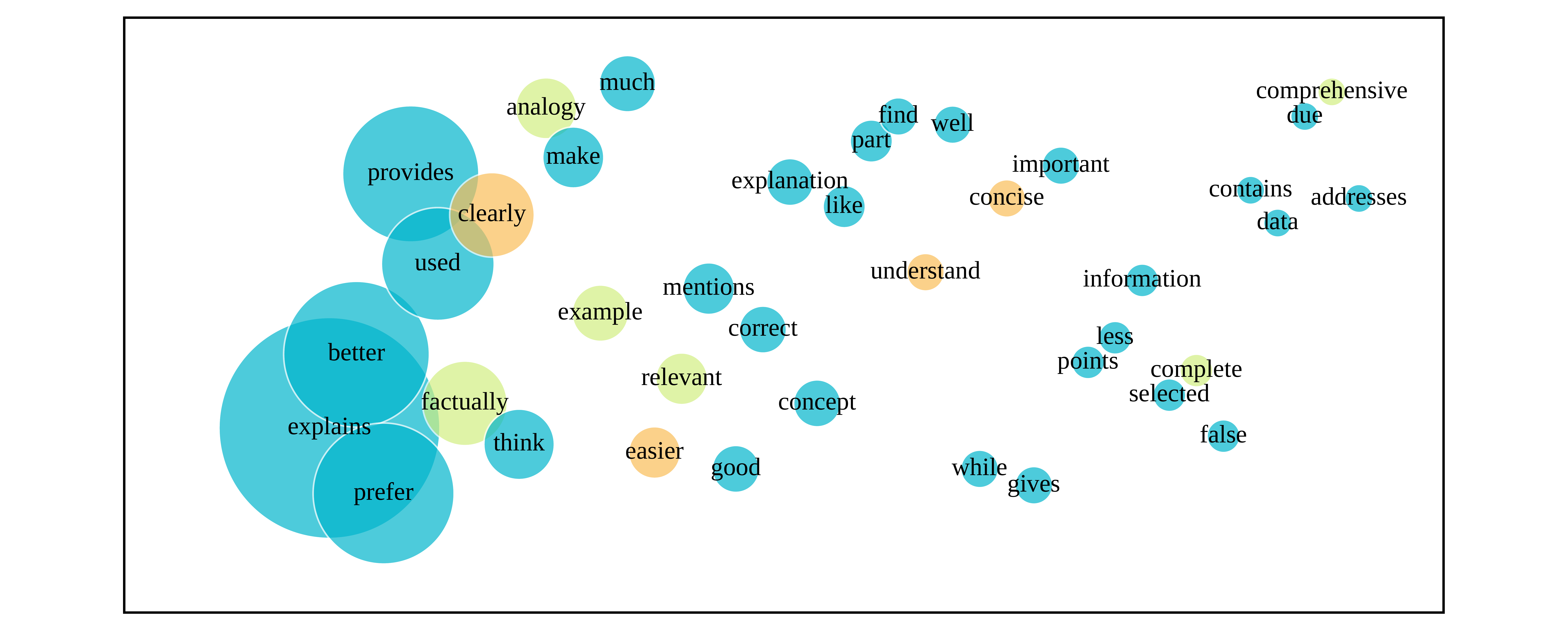}
\caption{
Distribution of the top 50 most common words mentioned by our expert annotators in their overall answer justifications. 
The size and color of the bubble represent the word frequency and importance, respectively. 
The green and orange colors denote the important evaluated and non-evaluated aspects, respectively, while blue depicts the generic terms used in answer justifications.}
\label{fig:ans_reason_frequency}
\end{figure*}

In \Cref{fig:ans_reason_frequency}, we plot the word frequency distribution of the free-form answer justifications provided by our expert annotators. Apart from our considered evaluation aspects, we observe that the annotators also find answers \textit{clarity}, \textit{conciseness}, and \textit{ease of understanding} helpful in deciding the overall best answer. We encourage future LFQA research to consider these aspects in their evaluation.

\subsection{Span-level Annotations}
\label{app:sec:span_level}
In \Cref{table:span_level_stats}, we present the distribution of errors annotated at different span-levels~(phrase, sentence and multi-sentence).
Our findings show that experts identify phrase-level errors in approximately $35\%$ of the cases, indicating that a substantial portion of errors are nuanced and cannot be effectively captured at the sentence level. This highlights the need to employ fine-grained span-level annotations to enhance the evaluation process, as they provide deeper insight into the nature and exact location of errors, ultimately leading to improved answer quality by targeting specific errors.

\renewcommand{\arraystretch}{1.2}
\begin{table}[!t]
\footnotesize
\centering
\begin{tabularx}{\linewidth}{@{}Xccc@{}}
\toprule
\textbf{Error Type} & \multicolumn{3}{c}{\textbf{Annotated Spans}} \\
\cmidrule(lr){2-4}
& \makecell{\textbf{Phrase-}\\\textbf{level}}  
& \makecell{\textbf{Sentence-}\\\textbf{level}} 
& \makecell{\textbf{Multi-sentence-}\\\textbf{level}} \\
\midrule
Question Misconception   & 38.89\% & 52.47\% & 8.64\% \\
Factuality & 42.40\% & 44.88\% & 12.72\% \\
Relevance  & 25.00\% & 39.13\% & 35.87\% \\
Completeness & 35.81\% & 34.63\% & 29.56\% \\
References & 31.24\% & 30.77\% & 38.00\% \\

\midrule
\textbf{Average} & 34.67\% & 40.38\%  & 24.96\% \\
\bottomrule
\end{tabularx}
\caption{
Overview of error types and the corresponding annotation distribution across phrase-level, sentence-level, and multi-sentence-level errors.
}
\label{table:span_level_stats}
\end{table}

\subsection{Expert (dis)agreement.}
\label{app:sec: exp_agreement}
In \Cref{table:quantitative_results}, we report Krippendorf's alpha~\cite{doi:10.1080/19312450709336664} as a measure of agreement for experts' overall answer preference. Our expert annotators achieve moderate agreement in technology, history, and law, fair agreement in biology and economics, and slight agreement in physics and chemistry.\footnote{Interpretation of agreement follows \citet{DBLP:conf/acl/WongPA20}} We emphasize that the disagreement between experts is \emph{not} a failure of our evaluation. Instead, it highlights the challenges of identifying fine-grained errors in answers, affecting overall preference. 
Moreover, prior work has similar findings for human disagreement in LFQA evaluation~\cite{xu-etal-2023-critical}.

\section{Prompts}
\label{appendix: prompts}
This section lists the prompts for data collection, training the error detection model, and refining answers using our Error-informed approach.

\subsection{Data Collection}
\label{app: prompts: gpt_prompt}

We prompt GPT-4 in a zero-shot manner to generate responses to questions asked on the Reddit platform, as shown in \Cref{lst:gpt4_zero_shot}. We use the default generation parameters in OpenAI API with temperature=0.1 and max\_tokens=1.5*(human\_answer\_length). We specifically instruct the model to generate a response of length similar to the corresponding human response on Reddit to compare model-generated and human-written answers fairly on our defined evaluation criteria.

\begin{listing}[!t]
\begin{minted}[fontsize=\footnotesize, frame=single, breaklines]{python}
f"""Your task is to answer a question by providing a clear and concise explanation of a complex concept in a way that is accessible for laypeople. The question was posted on the Reddit forum Explain Like I'm Five (r/explainlikeimfive). Please keep in mind that the question is not literally meant for 5-year-olds, so you should not answer the question in a way that you are talking to a child. Your answer should be around {human_answer_length} words and should break down the concept into understandable parts, providing relevant examples or analogies where appropriate. You should also aim to make your explanation easy to follow, using clear and concise language throughout. Your answer should maintain accuracy and clarity. When appropriate, you can start with one sentence summarizing the main idea of the answer.

Question: {question}   

Answer (around {human_answer_length} words):
"""
\end{minted}
\caption{Zero-shot prompt for GPT-4 to generate long-form answers to questions asked on the ELI5 subreddit on the reddit platform.}
\label{lst:gpt4_zero_shot}
\end{listing}

\subsection{Feedback Model}
\label{app: prompts: feedback}
We use expert error annotations for the \textit{completeness} aspect from our HQ$^2$A dataset and train the feedback model for 5 epochs with a learning rate $2e-5$ and a sequence length of 1024. 
In \Cref{lst: feedback_model_prompt}, we show an example prompt used to train our feedback model.
Given an instruction and input question-answer, the output is a sentence-level prediction of answer completeness with detailed justifications.

 There are clear patterns of how humans assess incomplete answers, which the model can learn from our training data. We explain them below, along with examples from our annotated dataset:

 \begin{enumerate}
     \item \textbf{Missing explanation of key concepts in answer.} These refer to concepts that are introduced in the answer but fail to explain it properly, assuming the reader already knows about it.

     \textit{Example}: 

     \textbf{Q}: How does rendering a video game resolution above your monitor resolution make the picture more crisp? You would think the resolution is already "maxed out", how does this improve things?
     
    \textbf{A}: It's called supersampling and basically gives the computer more options for what to render for any given pixel. If you "render" at a higher resolution so that each final pixel is 4 pixels in the supersample then the computer gets 4 reference points for what color that pixel could be instead of just one. [...]
    
    \textbf{Comment}: Supersampling is not explained clearly in this answer.

    \item \textbf{Failure to Address All Parts of the Question.} The answer answers only a part of the question.

     \textit{Example}: 

     \textbf{Q}: Why can’t spam call centers be automatically shut down? Additionally, why can’t spam calls be automatically blocked, and why is nobody really doing a whole lot about it? It seems like this is a problem that they would have come up with a solution for by now.

    \textbf{A}: Spam call centers can't be automatically shut down because they are often located in countries with weak regulation and they use advanced techniques to hide their identity, like caller id spoofing. Blocking spam calls is challenging because these callers use ever-changing phone numbers, making it difficult to maintain an updated block list. While there are some efforts to combat spam calls, such as telecom companies implementing call-filtering systems and governments enacting regulations, the issue is deeply rooted in the economic incentive for spam callers and the complexities in tracking them down. Ultimately, tackling spam calls requires international cooperation and continuous technology advancements.
    
    \textbf{Comment}:  This answer is incomplete because it does not fully answer parts of the question but only hints at them, such as “Why is nobody really doing a whole lot about it?” and “Why hasn’t a solution been found by now?”.

    \item \textbf{Unfinished sentences or the lack of logical flow.} Sometimes, a sentence in the answer may be left unfinished or lacks a clear, logical flow, making it hard to understand.

     \textit{Example}: 

     \textbf{Q}: How can a collision cause a loss of energy? Even if friction and energy loss is ignored?

    \textbf{A}: There isn't any energy loss. It's conserved. There is a kinetic energy loss because, in most collisions, there's heat, sound, and (often) distortion of the colliding objects. The energy changed forms to some that aren't kinetic, so naturally you've got less kinetic energy after than before. Your second equation (vf2=...) is just conservation of momentum. This doesn't looks like it's an energy question.
    
    \textbf{Comment}:  The first sentence says "There isn't any energy loss. It's conserved." But immediately after, the answer states "There is a kinetic energy loss..." without making a clear distinction between total energy conservation and kinetic energy loss.

    \item \textbf{Lack of examples to provide more clarity.} Incomplete answers may lack references or examples to provide clarity to the reader.

     \textit{Example}: 

     \textbf{Q}: How come a CPU can function without any driver updates whereas a GPU needs one every month or so for stability?

    \textbf{A}: The eli5 answer is that you do run updates for your CPU, just don’t notice that because they are packaged into stuff like iOS updates or windows updates. Operating systems have a thing called “kernel” that can be seen as the driver of your CPU (yes yes I know is not a perfect analogy but this is eli5) and that is updated regularly by windows/mac/linux operating systems.

    \textbf{Comment}: It doesn't mention any example of such an update or list any type of such update to give a clearer picture of the event it is describing.

    \item \textbf{Insufficient justification for a claim.} Ideas/claims without backing them up with evidence or explanations.

     \textit{Example}: 

     \textbf{Q}: If most electronic appliances' efficiency losses are through heat, does that mean that electric heaters are 100\% efficient?

    \textbf{A}: Yes, electric heaters convert 100\% of the power that they consume into heat. So they have an efficiency of 100\%. Heat pumps move heat from one area (outside your house) to another area (inside your house) the amount of heat they move is typically about 3 times more than the power they consume. So the in terms of energy-to-heat efficiency, they are 300\%+ efficient. But thermodynamically they are not “creating” heat from nothing. So heat pumps are not perpetual motion machines, they don’t break any of the laws of thermodynamics.

    \textbf{Comment}: The answer lacks sufficient justification for the claim that electric heaters are 100\% efficient. More information about the mechanisms it uses to achieve that 100\% efficiency would make the answer more complete.

 \end{enumerate}

\begin{listing}[!h]
\begin{minted}[fontsize=\footnotesize, frame=single, breaklines]{python}
f"""### Instruction:
When given a question and answer statements, evaluate whether each given statement provides sufficient information for answering the question. 
Use the '[Incomplete]' tag to indicate answer incompleteness, and '[Complete]' tag to indicate completeness, with reasons.
Please note that the answer can have single, multiple or no incomplete statements.

### Input:
Question: Can anyone explain the differences between copyright and trademark?
Answer: 1. A trademark protects a brand's symbol or logo.
2. A copyright protects content.
3. So the ac/dc logo with the lightning bolt would be trademarked.
4. The music and lyrics to thunderstruck would be copyrighted.
5. Edit: eli10 addendum: just to be clear, the content of a copyright can also be an image.
6. So the album cover to thunderstruck's album, razor's edge, would be copyrighted because it is artistic content owned by someone, but doesn't identify ac/dc as a whole.

### Response: 1. [Complete]
2. [Incomplete] Reasons: The answer fails to mention the broader scope of copyright protection, which includes creative works beyond just music and lyrics.
3. [Complete]
4. [Complete]
5. [Complete]
6. [Complete]
"""
\end{minted}
\caption{An example prompt used for training LLaMA2-13B model for error feedback.}
\label{lst: feedback_model_prompt}
\end{listing}

\subsection{Refinement Model}
\label{app: prompts: refinement}

As detailed in \Cref{sec: error-informed refinement}, the refinement model uses coarse-grained feedback~(\textsc{Improve} and \textsc{Generic}) and fine-grained feedback from the learned error detection model to refine input answers. We list the prompts used for \textsc{Improve}, \textsc{Generic} and incorporating fine-grained feedback in \Cref{lst:llama_no_feedback_prompt}, \Cref{lst:llama_generic_feedback_prompt} and \Cref{lst:llama_error_feedback_prompt}, respectively.

\begin{listing}[!h]
\begin{minted}[fontsize=\footnotesize, frame=single, breaklines]{python}
f"""
Answer the following question: "{question}"
Your answer is: "{answer}".
Please improve your answer.
Your improved answer:

"""
\end{minted}
\caption{Zero-shot prompt for LLaMA2-13B-chat model to refine long-form answers without feedback from the error detection model~(\textsc{Improve}).}
\label{lst:llama_no_feedback_prompt}
\end{listing}

\begin{listing}[!h]
\begin{minted}[fontsize=\footnotesize, frame=single, breaklines]{python}
f"""
Answer the following question: "{question}"
Your answer is: "{answer}".
The answer is not complete.
Please improve your answer.
Your improved answer:

"""
\end{minted}
\caption{Zero-shot prompt for LLaMA2-13B-chat model to refine long-form answers with generic feedback~(\textsc{Generic}).}
\label{lst:llama_generic_feedback_prompt}
\end{listing}

\begin{listing}[!t]
\begin{minted}[fontsize=\footnotesize, frame=single, breaklines]{python}
f"""
Answer the following question: "{question}"
Your answer is: "{answer}".
The answer is not complete because: 
"{reason}".
Please improve your answer.
Your improved answer:

"""

# reasons are given as:
# 1. Reason 1
# 2. Reason 2
# ...
\end{minted}
\caption{Zero-shot prompt for LLaMA2-13B-chat model to refine long-form answers with error feedback from the error detection model.}
\label{lst:llama_error_feedback_prompt}
\end{listing}

\section{Mitigating Errors with Preference Optimization}
\label{app: sec: dpo}

While language models acquire large amounts of world knowledge and strong reasoning skills from unsupervised training over massive web corpora, aligning them with human expectations is often hard. 
Model alignment techniques like direct preference optimization~(DPO)~\cite{DBLP:conf/nips/RafailovSMMEF23} allow us to directly use preference data to optimize the language model by casting the RL-based objective used by existing RLHF methods to an objective that can be directly optimized via a simple binary cross-entropy loss. This simplifies the process of refining LLMs greatly. The following paragraphs detail how we use DPO to reduce LLM errors.

\paragraph{Implementation details.} 
We model data from \textsc{HQ$^2$A} as a preference dataset where every question has a chosen and a rejected response selected by expert annotators based on the given evaluation criteria. Using this dataset, we fine-tune the LLaMA2-7B-chat~\cite{DBLP:journals/corr/abs-2307-09288} and Mistral-7B-Instruct-v0.1~\cite{DBLP:journals/corr/abs-2310-06825} models with the DPO algorithm. We use $batch\_size=16$, $warmup\_ratio=0.1$, $learning\_rate=2e-5$, $num\_epochs=5$, $beta=0.1$, and $max\_length=1024$ for training the models.

Due to compute limitations, we train Llama2-13B-chat model on our preference dataset using LoRA~\cite{DBLP:conf/iclr/HuSWALWWC22}. We use the following training parameters: $r=256$, $alpha=128$, $lora\_dropout=0.05$, $learning\_rate=5e-5$, $beta=0.1$, $max\_length=1024$ and train the model for 5 epochs.

\paragraph{Datasets \& Evaluation Metrics.}
\label{sec: datasets}
We experiment with three datasets: \textsc{HQ$^2$A}, ASQA~\cite{stelmakh-etal-2022-asqa}, and ELI5~\cite{fan-etal-2019-eli5}. \textsc{HQ$^2$A} dataset consists of 698 high-quality long-form question-answer pairs split into train~(80\%), dev~(10\%), and test~(10\%) sets. The ASQA dataset consists of 6K ambiguous factoid questions with long-form answers synthesized from multiple sources to resolve the ambiguities. ELI5 consists of 270K long-form answers covering general topics from the subreddits "explainlikeimfive", "askscience", and "AskHistorians" on the Reddit platform.

We report the quality of the generated long-form answers using TigerScore~\cite{jiang2023tigerscore}, a trained reference-free evaluation metric to pinpoint mistakes in the LLM-generated text. TigerScore detects errors in the input text and assigns an error score based on the severity of the error detected. Specifically, we use the LLaMA-7B trained version of TigerScore, which highly correlates with humans for error detection in LFQA tasks~\cite{jiang2023tigerscore}. We also measure the factual correctness of the generated answers using sample-based consistency metrics~\cite{DBLP:conf/emnlp/ManakulLG23}. Following their approach, we zero-shot prompt a LLaMA-13B-chat model to check if $i^{th}$ sentence in the original answer is supported by the sampled answer $S^n$ and return a score $x_i^n$ using the mapping: \{"Yes: 1.0", "No: 0.0", "N/A: 0.5"\}. The final consistency score is then calculated as:

\begin{equation*}
S_{Prompt} (i) = \frac{1}{N} \sum_{n=1}^N x_i^n
\end{equation*}

\section{Training, Infrastructure and Runtime}

We use a server with $8$ NVIDIA A$100$ Tensor Core GPUs, each with $80$GB VRAM, to run all our experiments.
Each experiment required, at most, two A$100$ GPUs.
Fine-tuning the LLaMA2-13B feedback model took $6$ hours on 2 A$100$ GPUs using our \textsc{HQ$^2$A} dataset.
LoRA fine-tuning of the LLaMA2-13B-chat refinement model took $2$ hours on a single A$100$ GPU using the preference data from \textsc{HQ$^2$A}.
Refining answers with our \textsc{Error-Informed Refinement} approach took $0.5$, $3$, and $23$ hours for the \textsc{HQ$^2$A}, ASQA, and ELI5 datasets, respectively, on a single A$100$ GPU.
The evaluation of the refined answers with TigerScore~(LLaMA-7B) utilized the VLLM inference library~\cite{kwon2023efficient} and took approximately $1$, $15$, and $30$ minutes for \textsc{HQ$^2$A}, ASQA, and ELI5 datasets, respectively, on a single A$100$ GPU.

\section{Additional Results}

\subsection{Aligning LLMs}
\label{app: align_lms}

\Cref{table:dpo_preference} shows the results for training language models with DPO using our collected preference annotations. Our preference-tuned models outperform the strong baseline models and reduce error generations in all the evaluation settings except the LLaMA model on the ASQA dataset. We hypothesize that this is due to the ambiguous nature of questions in the ASQA dataset that can have multiple correct answers.

We also observe that the models become more robust and generate more consistent responses after preference-tuning. The only exception is the Mistral model on our held-out test set, which has lower response consistency. We believe this is likely due to the conservative nature of DPO-trained models wherein, during sampling, it can refrain from answering a question in some cases and not in others, leading to a lower consistency score.

\renewcommand{\arraystretch}{1.2}
\begin{table*}[!t]
\notsotiny
\centering
\begin{tabularx}{\textwidth}{@{}XXlll@{}}
\toprule
\multirow{2.5}{1.5cm}{\textbf{Dataset} \\ \textbf{(\# samples)}} & \multirow{2.5}{3cm}{\textbf{Instruct Model}} & \multicolumn{2}{c}{\textbf{TIGERScore}} & \multirow{2.5}{3cm}{\textbf{SelfCheck} \\ \textbf{Consistency ($\downarrow$)}}\\
\cmidrule(lr){3-4}
& & \% Error samples ($\downarrow$)  & Error score ($\downarrow$) \\
\midrule
\multirow{4}{2.5cm}{\textsc{HQ$^2$A}~(70)} & LLaMA2-7B & $18.57\pm0.00$ &  $\textbf{0.60}\pm0.00$ & $0.166\pm0.014$ \\
& LLaMA2-7B + DPO  & $\textbf{15.71}\pm0.00$  & $0.66\pm0.00$ &  $\textbf{0.162}\pm0.015$  \\
\cline{2-5}
& Mistral-7B  & $20.00\pm0.00$ & $0.57\pm0.00$ & $\textbf{0.266}\pm0.011$ \\
& Mistral-7B + DPO  & $\textbf{17.14}\pm0.00$ & $\textbf{0.54}\pm0.00$ &  $0.285\pm0.011$ \\
\midrule

\multirow{4}{2.5cm}{\textsc{ASQA}~(948)} & LLaMA2-7B & $\textbf{26.58}\pm1.49$  & $\textbf{0.86}\pm0.06$ & $0.187\pm0.014$ \\
& LLaMA2-7B + DPO  & $28.41\pm1.06$ & $0.89\pm0.02$ & $\textbf{0.178}\pm0.006$ \\
\cline{2-5}
& Mistral-7B & $62.09\pm0.35$ & $2.08\pm0.01$ & $0.578\pm0.003$ \\
& Mistral-7B + DPO  & $\textbf{60.80}\pm0.56$  & $\textbf{2.03}\pm0.01$ &  $\textbf{0.555}\pm0.008$  \\
\midrule

\multirow{4}{2.5cm}{\textsc{ELI5\_general}\\(1000)} & LLaMA2-7B & $9.93\pm1.05$  & $0.32\pm0.04$ & $0.133\pm0.001$ \\
& LLaMA2-7B + DPO  & $\textbf{9.33}\pm0.66$  & $\textbf{0.29}\pm0.03$ & $\textbf{0.130}\pm0.004$  \\
\cdashline{2-5}
& Mistral-7B  & $29.97\pm0.97$ & $0.90\pm0.04$ & $0.327\pm0.003$ \\
& Mistral-7B + DPO  & $\textbf{22.77}\pm1.03$  & $\textbf{0.72}\pm0.03$ & $\textbf{0.319}\pm0.011$ \\
\midrule

\multirow{4}{2.5cm}{\textsc{ELI5\_science}\\(1000)} & LLaMA2-7B & $\textbf{9.47}\pm0.47$  & $0.31\pm0.02$ & $\textbf{0.137}\pm0.003$ \\
& LLaMA2-7B + DPO  & $\textbf{9.47}\pm0.76$  & $\textbf{0.30}\pm0.00$ & $0.139\pm0.004$ \\
\cdashline{2-5}
& Mistral-7B  & $34.10\pm0.94$  & $1.07\pm0.02$ &  $0.320\pm0.004$ \\
& Mistral-7B + DPO  & $\textbf{29.03}\pm1.51$  & $\textbf{0.95}\pm0.04$ & $\textbf{0.297}\pm0.010$ \\
\midrule

\multirow{4}{2.5cm}{\textsc{ELI5\_history} \\ (1000)} & LLaMA2-7B & $9.63\pm0.59$  &  $0.30\pm0.02$ &  $\textbf{0.188}\pm0.005$ \\
& LLaMA2-7B + DPO  & $\textbf{7.60}\pm0.08$  & $\textbf{0.22}\pm0.01$ & $0.189\pm0.005$ \\
\cdashline{2-5}
& Mistral-7B  & $26.23\pm0.38$  & $0.79\pm0.02$ & $0.363\pm0.016$ \\
& Mistral-7B + DPO  & $\textbf{22.17}\pm1.31$ & $\textbf{0.69}\pm0.04$ & $\textbf{0.345}\pm0.013$  \\

\bottomrule
\end{tabularx}
\caption{Results of aligning LLMs with DPO using our collected answer preference data. We measure the errors using Tigerscore and the consistency of model outputs using SelfCheckGPT. Reported results are averages over three iterations with standard deviations. The best scores are marked in \textbf{bold}.
}
\label{table:dpo_preference}
\end{table*}

\subsection{EIR with DPO}
\label{app: sec: eir_dpo}
In \Cref{table:feedback_refine_dpo}, we present the quality of answers refined using different types of feedback (coarse- and fine-grained), alongside the \textsc{baseline} answers. Additionally, we include the results for answers refined with the DPO-aligned model. While the DPO-aligned refinement model does not outperform the vanilla refinement model in reducing the overall number of error samples, it achieves the best error scores on ASQA and ELI5. This suggests that the DPO optimization is still effective in correcting major errors to some extent.

\begin{table*}[htbp]
\notsotiny
\centering
\begin{tabularx}{\textwidth}{@{}llllXXX@{}}
\toprule
\multirow{3}{*}{\textbf{Dataset}} & \multirow{3}{*}{\textbf{Approach}} & \multicolumn{2}{c}{\textbf{TIGERScore}} & \multicolumn{3}{c}{\textbf{Error Correction}} \\
\cmidrule(lr){3-4} \cmidrule(lr){5-7} 
& & \multicolumn{1}{l}{\textbf{\% Error samples ($\downarrow$)}} & \multicolumn{1}{l}{\textbf{Error score ($\downarrow$)}} & \multicolumn{1}{l}{\textbf{Precision ($\uparrow$)}} & \multicolumn{1}{l}{\textbf{Recall ($\uparrow$)}} & \multicolumn{1}{l}{\textbf{F1 ($\uparrow$)}} \\

\midrule

\multirow{2}{1.5cm}{\textsc{HQ$^2$A}}  & Human feedback & $2.61\pm0.92$  & $0.09\pm0.01$ & $0.86\pm0.04$ &  $\textbf{1.00}\pm0.00$ & $0.94\pm0.02$  \\
\cdashline{2-7}
& Baseline &  $19.61$  & $0.63$ & - & - & -  \\
 & Improve & $1.31\pm0.92$  & $0.05\pm0.04$ & $\textbf{1.00}\pm0.00$ & $0.93\pm0.05$ & $0.97\pm0.02$ \\
 & Generic & $1.31\pm0.92$  & $0.05\pm0.03$ & $0.97\pm0.04$  & $0.97\pm0.05$  & $0.97\pm0.02$ \\
 \cdashline{2-7}
 & \textsc{EIR} (\textit{Ours}) & $\textbf{0.65}\pm0.92$  & $\textbf{0.03}\pm0.04$ & $0.97\pm0.04$ &  $\textbf{1.00}\pm0.00$ & $\textbf{0.98}\pm0.02$  \\
 & \textsc{EIR} w/ DPO (\textit{Ours}) & $4.57\pm2.44$  & $0.07\pm0.02$ & $0.90\pm0.08$ &  $0.87\pm0.05$ & $0.88\pm0.06$  \\

\midrule

\multirow{2}{1.5cm}{\textsc{ASQA}}  &  Baseline & $34.81$ & $1.20$ & - & - & - \\
 & Improve & $20.85\pm1.00$  & $0.68\pm0.03$ & $0.70\pm0.02$ &  $0.71\pm0.01$ & $0.70\pm0.01$ \\
 & Generic &  $18.67\pm0.52$ & $0.61\pm0.01$ & $0.72\pm0.01$ &  $0.75\pm0.01$ & $0.74\pm0.00$ \\
 \cdashline{2-7}
 & \textsc{EIR} (\textit{Ours}) &  $\textbf{16.63}\pm0.41$  & $0.51\pm0.02$ & $\textbf{0.73}\pm0.00$ & $\textbf{0.82}\pm0.02$ & $\textbf{0.77}\pm0.01$ \\
 & \textsc{EIR} w/ DPO (\textit{Ours}) & $22.61\pm0.26$  & $\textbf{0.45}\pm0.01$ & $0.64\pm0.00$ &  $0.77\pm0.01$ & $0.71\pm0.00$  \\

\midrule

\multirow{2}{1.5cm}{\textsc{ELI5}}  &  Baseline & $22.93$   & $0.82$ & - &  - & -  \\
 & Improve & $10.05\pm0.18$  & $0.36\pm0.02$  & $0.75\pm0.00$  & $0.86\pm0.00$ & $0.80\pm0.00$  \\
 & Generic & $6.06\pm0.23$  & $0.22\pm0.01$  & $0.84\pm0.01$ & $0.91\pm0.00$ & $0.87\pm0.00$ \\
 \cdashline{2-7}
 & \textsc{EIR} (\textit{Ours}) &  $\textbf{3.81}\pm0.30$ &  $\textbf{0.13}\pm0.01$ & $\textbf{0.88}\pm0.01$   &  $\textbf{0.96}\pm0.01$  & $\textbf{0.92}\pm0.01$  \\
 & \textsc{EIR} w/ DPO (\textit{Ours}) & $5.71\pm0.25$  & $\textbf{0.13}\pm0.00$ & $0.83\pm0.00$ &  $0.94\pm0.01$ & $0.88\pm0.00$  \\

\bottomrule
\end{tabularx}
\caption{
Results on the quality of original answers from the datasets~(\textsc{Baseline}), answers refined with coarse-grained feedback~(\textsc{Improve} and \textsc{Generic}), and fine-grained feedback~(\textsc{EIR}). Additionally, we include the results of refinement with expert human feedback on our collected data. Reported results are averages over three iterations with standard deviations. The best scores are marked in \textbf{bold}.
}
\label{table:feedback_refine_dpo}
\end{table*}

\subsection{Fine- vs. Coarse-grained Feedback}
\label{app: additional_refinement_models}

In \Cref{table:feedback_refine_llama3} and \Cref{table:feedback_refine_mistral}, we show the results on the quality of answers generated with zero-shot prompting~(\textsc{Zero-shot}) as well as answers refined using coarse~(\textsc{Improve} and \textsc{Generic}) and fine-grained~(\textsc{EIR}) feedback, using the LLaMA3-8B-Instruct and Mistral-7B-Instruct-v0.3 models, respectively. Similar to the observations in \Cref{sec: results_feedback}, we notice that inadequate feedback deteriorates the quality of generation. 

When using LLaMA3-8B-Instruct as the refinement model, the direct prompting~(\textsc{Zero-shot}) and refining without detailed feedback~(\textsc{Improve}) approaches improve answer quality over the \textsc{Baseline}~(original answers from the dataset) on all the datasets, except ASQA, where the \textsc{Zero-shot} approach generates lower quality answers than \textsc{Baseline}, likely due to the ambiguous nature of the questions in the ASQA dataset. On the contrary, prompting with more targeted feedback (\textsc{Generic}) consistently outperforms the \textsc{Baseline}, \textsc{Zero-shot}, and \textsc{Improve} approaches, generating better quality LFQA answers and giving the best scores on \textsc{HQ$^2$A}.
Furthermore, providing fine-grained feedback from our error detection model~(\textsc{EIR}) outperforms coarse-grained feedback on ASQA and ELI5 datasets, reducing error samples and error scores by \textasciitilde$8\%$ and  \textasciitilde$\Delta{68}\%$, respectively, and improving F1 scores by \textasciitilde$11\%$ on average.

When using Mistral-7B-Instruct-v0.3 as the refinement model, the approach to refine answers without detailed feedback~(\textsc{Improve}) improves answer quality over the \textsc{Baseline}, \textsc{Zero-shot}, and even the \textsc{Generic} approach, achieving the best scores on \textsc{HQ$^2$A}. We hypothesize that this is due to the capability of the model to understand simplistic feedback instructions to improve answers, leading to a better performance than the \textsc{Generic} approach.
In contrast, providing fine-grained feedback from our error detection model~(\textsc{EIR}) outperforms coarse-grained feedback on ASQA and ELI5 datasets, reducing error samples and error scores by \textasciitilde$2\%$ and  \textasciitilde$\Delta{27}\%$, respectively, and improving F1 scores by \textasciitilde$4\%$ on average.

\begin{table*}[htbp]
\notsotiny
\centering
\begin{tabularx}{\textwidth}{@{}llllXXX@{}}
\toprule
\multirow{3}{*}{\textbf{Dataset}} & \multirow{3}{*}{\textbf{Approach}} & \multicolumn{2}{c}{\textbf{TIGERScore}} & \multicolumn{3}{c}{\textbf{Error Correction}} \\
\cmidrule(lr){3-4} \cmidrule(lr){5-7} 
& & \multicolumn{1}{l}{\textbf{\% Error samples ($\downarrow$)}} & \multicolumn{1}{l}{\textbf{Error score ($\downarrow$)}} & \multicolumn{1}{l}{\textbf{Precision ($\uparrow$)}} & \multicolumn{1}{l}{\textbf{Recall ($\uparrow$)}} & \multicolumn{1}{l}{\textbf{F1 ($\uparrow$)}} \\

\midrule

\multirow{2}{1.5cm}{\textsc{HQ$^2$A}}  
& Human feedback  & $1.96\pm0.00$  & $0.07\pm0.01$ & $0.90\pm0.01$   & $\textbf{1.00}\pm0.00$   & $0.95\pm0.00$ \\
\cdashline{2-7}
& Baseline &  $19.61$  & $0.63$ & - & - & -  \\
& Zero-shot$_{\text{LLaMA3}}$ &  $17.65\pm0.00$  & $0.46\pm0.00$ & $0.53\pm0.00$ & $0.80\pm0.00$ & $0.64\pm0.00$  \\
 &  Improve  & $2.61\pm0.92$  & $0.04\pm0.04$ &  $0.88\pm0.04$  &  \greenbox{$\textbf{1.00}\pm0.00$}  & $0.93\pm0.02$ \\
 &  Generic  & \greenbox{$\textbf{0.00}\pm0.00$}  &\greenbox{$\textbf{0.00}\pm0.00$} &  \greenbox{$\textbf{1.00}\pm0.00$}  &  \greenbox{$\textbf{1.00}\pm0.00$}  & \greenbox{$\textbf{1.00}\pm0.00$}  \\
 \cdashline{2-7}
 &  \textsc{EIR} (\textit{Ours})  & \orangebox{$1.30\pm0.92$}  & \orangebox{$0.03\pm0.03$} &  \orangebox{$0.96\pm0.04$}  &  \orangebox{$0.96\pm0.05$}  & \orangebox{$0.96\pm0.02$} \\ 
   
\midrule

\multirow{2}{1.5cm}{\textsc{ASQA}}  &  Baseline & $34.81$ & $1.20$ & - & - & - \\
& Zero-shot$_{\text{LLaMA3}}$ &  $42.83\pm0.00$ & $1.39\pm0.00$ & $0.41\pm0.00$ & $0.55\pm0.00$ & $0.47\pm0.00$  \\
 &  Improve  & $30.09\pm0.53$  & $0.82\pm0.01$ & $0.55\pm0.01$   & $0.72\pm0.01$   & $0.62\pm0.01$ \\
 &  Generic  &  \orangebox{$20.92\pm0.62$} & \orangebox{$0.51\pm0.03$}  & \orangebox{$0.66\pm0.01$}   &  \orangebox{$0.81\pm0.01$}  & \orangebox{$0.72\pm0.01$} \\
 \cdashline{2-7}
 &  \textsc{EIR} (\textit{Ours})  &  \greenbox{$\textbf{10.16}\pm0.65$} & \greenbox{$\textbf{0.23}\pm0.02$}  & \greenbox{$\textbf{0.82}\pm0.02$}   &  \greenbox{$\textbf{0.89}\pm0.01$}  &  \greenbox{$\textbf{0.85}\pm0.01$}\\

\midrule

\multirow{2}{1.5cm}{\textsc{ELI5}}  &  Baseline & $22.93$   & $0.82$ & - &  - & -  \\
& Zero-shot$_{\text{LLaMA3}}$ &  $3.22\pm0.00$  & $0.10\pm0.00$ & \orangebox{$0.91\pm0.00$} & $0.96\pm0.00$ & $0.93\pm0.00$  \\
 &  Improve  & $3.05\pm0.14$  & $0.09\pm0.01$ &  $0.90\pm0.01$  & \orangebox{$0.97\pm0.00$}   & $0.93\pm0.01$ \\
 &  Generic  & \orangebox{$2.70\pm0.18$}  & \orangebox{$0.06\pm0.01$} & $0.91\pm0.01$   & \orangebox{$0.97\pm0.00$}   & \orangebox{$0.94\pm0.00$} \\
 \cdashline{2-7}
 &  \textsc{EIR} (\textit{Ours}) &  \greenbox{$\textbf{0.99}\pm0.06$} & \greenbox{$\textbf{0.02}\pm0.01$} & \greenbox{$\textbf{0.96}\pm0.01$}   & \greenbox{$\textbf{0.99}\pm0.00$}   &  \greenbox{$\textbf{0.97}\pm0.01$} \\

\bottomrule
\end{tabularx}
\caption{
Results on the quality of original answers from the datasets~(\textsc{Baseline}), answers from 0-shot prompting LLaMA3-8B-Instruct~(\textsc{Zero-shot}), answers refined with coarse-grained feedback~(\textsc{Improve} and \textsc{Generic}), and fine-grained feedback~(\textsc{EIR}) using LLaMA3-8B-Instruct refinement model. Additionally, we include the results of refinement with expert human feedback on our collected data. Reported results are averages over three iterations with standard deviations. he best results are in \greenbox{\textbf{bold green}} and the second-best results are in \orangebox{orange}.
}
\label{table:feedback_refine_llama3}
\end{table*}


\begin{table*}[htbp]
\notsotiny
\centering
\begin{tabularx}{\textwidth}{@{}llllXXX@{}}
\toprule
\multirow{3}{*}{\textbf{Dataset}} & \multirow{3}{*}{\textbf{Approach}} & \multicolumn{2}{c}{\textbf{TIGERScore}} & \multicolumn{3}{c}{\textbf{Error Correction}} \\
\cmidrule(lr){3-4} \cmidrule(lr){5-7} 
& & \multicolumn{1}{l}{\textbf{\% Error samples ($\downarrow$)}} & \multicolumn{1}{l}{\textbf{Error score ($\downarrow$)}} & \multicolumn{1}{l}{\textbf{Precision ($\uparrow$)}} & \multicolumn{1}{l}{\textbf{Recall ($\uparrow$)}} & \multicolumn{1}{l}{\textbf{F1 ($\uparrow$)}} \\

\midrule

\multirow{2}{1.5cm}{\textsc{HQ$^2$A}}  
& Human feedback   & $1.96\pm0.00$  & $0.07\pm0.01$ &  \orangebox{$0.90\pm0.01$}   & \greenbox{$\textbf{1.00}\pm0.00$}   & \orangebox{$0.95\pm0.00$} \\
\cdashline{2-7}
& Baseline &  $19.61$  & $0.63$ & - & - & -  \\

& Zero-shot$_{\text{Mistral}}$ &  $3.92\pm0.00$  & $0.16\pm0.00$ & $0.83\pm0.00$ & \greenbox{$\textbf{1.00}\pm0.00$} & $0.91\pm0.00$  \\

 &  Improve   &  \greenbox{$\textbf{1.30}\pm1.85$} & \greenbox{$\textbf{0.03}\pm0.05$} &  \greenbox{$\textbf{0.96}\pm0.05$}  & \orangebox{$0.96\pm0.05$}   & \greenbox{$\textbf{0.96}\pm0.05$}  \\
 
 & Generic   &  \orangebox{$1.96\pm0.00$} & \orangebox{$0.05\pm0.03$} & \orangebox{$0.90\pm0.01$}   &  \greenbox{$\textbf{1.00}\pm0.00$}  & \orangebox{$0.95\pm0.00$}  \\
 \cdashline{2-7}
  &  \textsc{EIR} (\textit{Ours})   &  $4.57\pm1.85$ & $0.15\pm0.05$ &  $0.85\pm0.06$  &  $0.93\pm0.09$  &  $0.88\pm0.05$ \\

\midrule

\multirow{2}{1.5cm}{\textsc{ASQA}} 
&  Baseline & $34.81$ & $1.20$ & - & - & - \\

& Zero-shot$_{\text{Mistral}}$ &  $39.35\pm0.00$  & $1.24\pm0.00$ & $0.45\pm0.00$ & $0.58\pm0.00$ & $0.51\pm0.00$  \\

 &  Improve   &  \orangebox{$13.53\pm0.44$} & \orangebox{$0.32\pm0.03$} & \orangebox{$0.77\pm0.01$}   & \orangebox{$0.86\pm0.01$}   & \orangebox{$0.81\pm0.01$}   \\
 
 &  Generic   &  $15.85\pm1.05$ & $0.40\pm0.03$ &  $0.74\pm0.02$  &  $0.83\pm0.01$  & $0.78\pm0.01$   \\
 
 \cdashline{2-7}
 &  \textsc{EIR} (\textit{Ours})  &  \greenbox{$\textbf{10.72}\pm0.96$} & \greenbox{$\textbf{0.23}\pm0.02$} & \greenbox{$\textbf{0.81}\pm0.02$}   & \greenbox{$\textbf{0.90}\pm0.01$}    &  \greenbox{$\textbf{0.85}\pm0.01$} \\

\midrule

\multirow{2}{1.5cm}{\textsc{ELI5}}  
&  Baseline & $22.93$   & $0.82$ & - &  - & -  \\

& Zero-shot$_{\text{Mistral}}$ &  $7.91\pm0.00$  & $0.25\pm0.00$ & $0.79\pm0.00$ & $0.90\pm0.00$ & $0.84\pm0.00$  \\

 &  Improve   &  \orangebox{$3.22\pm0.16$} & \orangebox{$0.09\pm0.01$} & \orangebox{$0.89\pm0.01$}   &  \orangebox{$0.96\pm0.01$}  & \orangebox{$0.93\pm0.00$}  \\
 
 & Generic   &  $3.81\pm0.15$ & $0.11\pm0.01$ & $0.88\pm0.01$   &  \orangebox{$0.96\pm0.01$}  & $0.92\pm0.00$  \\
 
 \cdashline{2-7}
 & \textsc{EIR} (\textit{Ours}) & \greenbox{$\textbf{3.02}\pm0.19$}  & \greenbox{$\textbf{0.08}\pm0.01$} &  \greenbox{$\textbf{0.90}\pm0.01$}  &  \greenbox{$\textbf{0.97}\pm0.01$}  & \greenbox{$\textbf{0.94}\pm0.00$} \\

\bottomrule
\end{tabularx}
\caption{
Results on the quality of original answers from the datasets~(\textsc{Baseline}), answers from 0-shot prompting Mistral-7B-Instruct-v0.3~(\textsc{Zero-shot}), answers refined with coarse-grained feedback~(\textsc{Improve} and \textsc{Generic}), and fine-grained feedback~(\textsc{EIR}) using Mistral-7B-Instruct-v0.3 refinement model. Additionally, we include the results of refinement with expert human feedback on our collected data. Reported results are averages over three iterations with standard deviations. he best results are in \greenbox{\textbf{bold green}} and the second-best results are in \orangebox{orange}.
}
\label{table:feedback_refine_mistral}
\end{table*}

\subsection{Human Evaluation}
\label{app: human_eval}
This section presents additional details of our human evaluation of the answers refined with our Error-informed feedback approach. 
In \Cref{table:human_eval_agreement}, we present the agreement of our annotators on two evaluation metrics: comprehensiveness and overall answer preference. The annotators strongly agree that the refined answers are comprehensive, i.e., the answer contains all the required information as asked by the question. For the overall answer preference compared to the baseline, we observe weak agreement between annotators, primarily due to the low agreement value on the ASQA dataset. We hypothesize that the annotators struggle to align on ASQA due to the ambiguous nature of the questions in this dataset, which may have multiple correct answers, and choosing between two answers is difficult.

\begin{table}[!t]
\small
\centering
\begin{tabularx}{0.48\textwidth}{@{}>{\raggedright\arraybackslash}p{0.20\linewidth}p{0.40\linewidth}p{0.35\linewidth}@{}}
\toprule
\textbf{Dataset} & \textbf{Comprehensiveness} ($\uparrow$) & \textbf{Preference} ($\uparrow$) \\

\midrule

\textsc{HQ$^2$A} & 0.70 & 0.31  \\

\textsc{ASQA}  & 0.86 & 0.02  \\

\textsc{ELI5} & 0.92 & 0.61  \\

\midrule
\textbf{Average} & 0.83 & 0.31 \\

\bottomrule
\end{tabularx}
\caption{Agreement of annotators on the comprehensiveness and preference of refined answers over the baseline answers from three datasets.}
\label{table:human_eval_agreement}
\end{table}

\begin{table*}[!t]
\scriptsize
\centering
\begin{tabularx}{\textwidth}{p{1.2cm}p{2.3cm}XXX}
\toprule
\textbf{Aspect} \textbf{Question} & \textbf{Answer} & \textbf{Original Annotation} & \textbf{Expert Justification}  &  \textbf{Sentence-level Annotation} \\
\midrule
\multirow{4}{*}{{\parbox{1.8cm}{Completeness}}}  & What are the pros/cons of borrowing money from the bank vs paying cash when buying a house? And why would it ever be a good idea to borrow if you can pay cash and pay no interest to the bank?

& \hl{The main advantage of borrowing money from a bank (a mortgage) to buy a house is that it allows you to purchase a property without having all the cash up front. This enables people to become homeowners earlier in life and potentially benefit from the appreciation of the home’s value. Paying a mortgage also helps build credit history, which can be beneficial for obtaining future loans or credit cards. On the other hand, paying cash for a house has the advantage of no interest costs, which can save a substantial amount of money over time. Cash buyers may also have more negotiating power, as they are not reliant on bank approval and can close the deal quickly. In certain situations, it can be better to borrow money even if you have the cash. This is because investing the cash into diversified investments (stocks, bonds, etc. ) may provide better returns than paying off the mortgage early, although it comes with some level of risk. Additionally, having cash reserves can provide financial flexibility for emergencies or future opportunities.}

& \begin{minipage}[t]{3.5cm}This answer is missing the disadvantages of getting a mortgage and paying in cash. It could mention how getting a mortgage means you will have to pay interest, as well as any mortgage fees (such as arrangement fees or valuation fees). It could also discuss how your home could be put at risk if you are unable to keep up with the repayments since a mortgage is secured against the value of your property. In terms of paying in cash, the answer could explain that this will narrow your investment portfolio, thereby missing the opportunity to invest in other assets that could increase your wealth over time.\\Source: (Heming,2022) https://www.moneysupermarket.com/\\mortgages/advantages-and-disadvantages/
\end{minipage}

&  \begin{minipage}[t]{3.5cm}
    1. [Incomplete] Reasons: This answer is missing [...]\\
    2. [Incomplete] Reasons: This answer is missing [...]\\
      \[...\] \\
    8. [Incomplete] Reasons: This answer is missing [...]\\
    \end{minipage} \\

\midrule

\multirow{4}{*}{{\parbox{1.8cm}{Completeness}}}  & Why do people often say diet sodas are just as bad or worse than regular sodas?

& Diet sodas are often considered just as bad or worse than regular sodas because they contain artificial sweeteners instead of sugar. These sweeteners, like aspartame or sucralose, can have negative effects on the body. For example, they may confuse the body's natural ability to regulate calorie intake by tricking it into thinking it's consuming real sugar. This can lead to overeating and weight gain. Additionally, some studies have suggested that artificial sweeteners might negatively impact gut bacteria, potentially leading to health issues. While diet sodas have fewer calories than regular sodas, \hl{they can still contribute to poor overall health due to these artificial ingredients}. Therefore, many people believe they are no better than, or even worse than, regular sodas.

& \begin{minipage}[t]{3.5cm}
Overall health is a combination of nutrition and exercise, artificial ingredients do not necessarily contribute to poor health.
\end{minipage}

&  \begin{minipage}[t]{3.5cm}
    1. [Complete] \\
    2. [Complete] \\
      \[...\] \\
    6. [Incomplete] Reasons: Overall health is a [...] \\
    7. [Complete] \\
    \end{minipage} \\

\bottomrule
\end{tabularx}
\caption{Examples of transforming expert annotations for completeness to sentence-level annotations for training the feedback model.}
\label{table: feedback_data_train}
\end{table*}

\onecolumn
\section{Annotation Guidelines}
\label{app: sec: annotation_guidelines}

We have previously described our data collection setup in \Cref{sec: data_collection}.
This section provides additional details on the annotation interface, detailed task instructions, and annotation procedure.

\subsection{Annotation Interface}
\begin{figure*}[h]
  \centering
  \includegraphics[width=\textwidth]{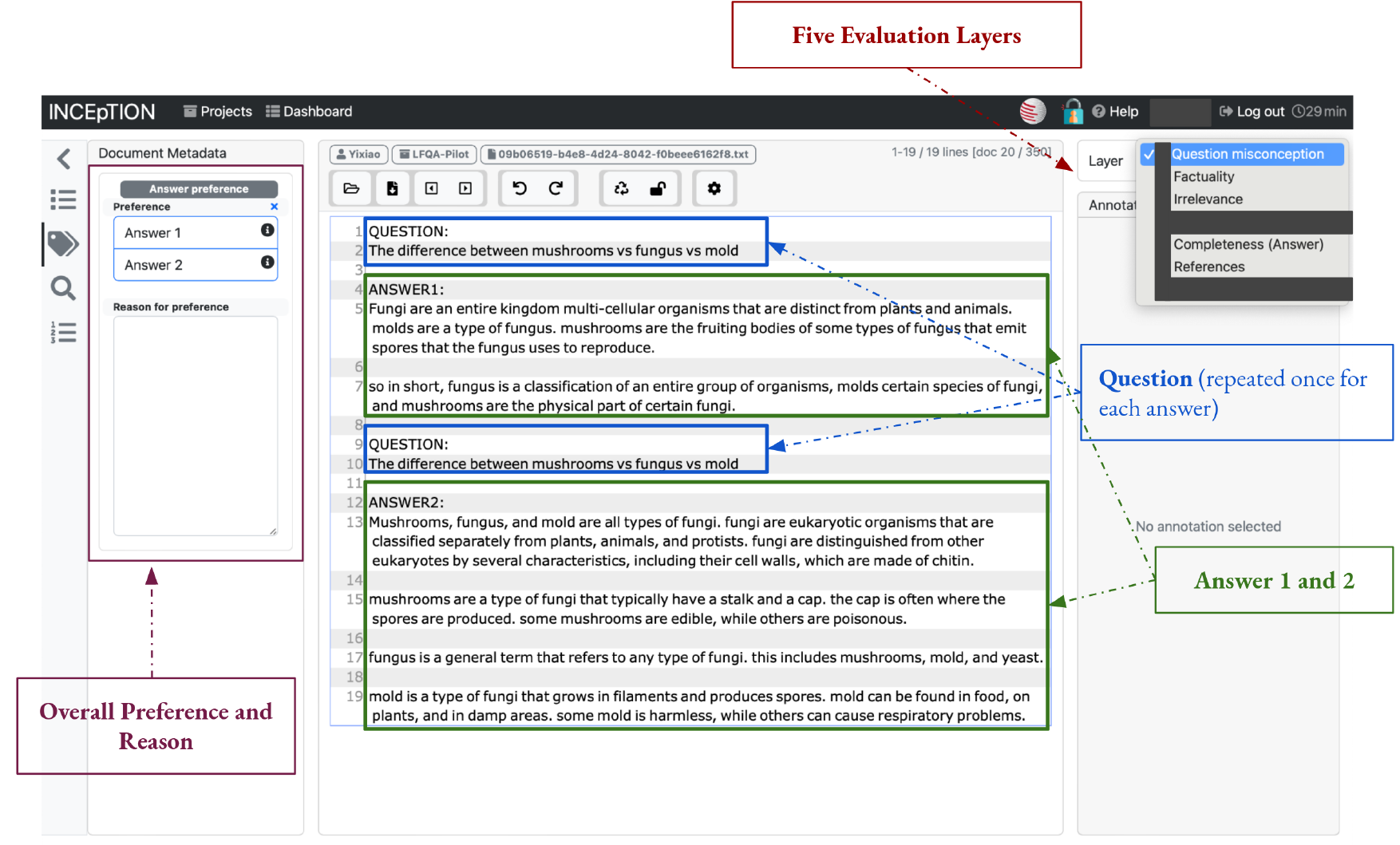}
  \caption{Screenshot of annotation interface for collecting expert error annotations on LFQA answers.}
  \label{fig:inception_platform}
\end{figure*}

In \Cref{fig:inception_platform}, we show the interface for collecting expert error annotations on LFQA answers. For every question, experts see a human-written and model-generated answer~(randomized order). Our expert annotators must select the evaluation layer~(top right) and highlight the error span in the question or answer, giving justifications with web references, wherever applicable. After annotating for all the evaluation criteria, experts judge the better answer and mark it in the left pane, giving reasons for their preference.

\subsection{Task Instructions}
\label{app:subsec: task_instructions}
We provide experts with detailed task instructions for evaluating answers according to the defined evaluation criteria. We go through every evaluation aspect in depth, defining it and giving annotation examples for clarification, as detailed in the next paragraphs.

\paragraph{1) Question Misconception.}

You should select a span of text in the \ul{question} that \textcolor{red}{contains a misconception or false assumption}. The question is repeated twice. You only need to select the span in one repetition. If you select such spans, we would like you to indicate in your reason (obligatorily): 

    \begin{itemize}
        \item whether the answers reject or correct the misconception/false assumption,

        \item if no answer rejects/corrects it, please explain in your reason why that is a misconception/false assumption (preferably with references).
    \end{itemize}

\noindent \textbf{Example:}

\noindent\emph{\ul{Question:}} 
Why is it so important for humans to have a balanced nutrition \hl{but not for animals}? Most animals have a fairly simple diet, \hl{carnivores eat only meat their whole life, cows eat exclusively grass etc.} So why are human bodies so picky and need a balance of protein, fat, carbs etc from different sources to perform well?

\paragraph{2) Factuality.}

You should select a span of text in the \ul{answers} that is \textcolor{red}{factually incorrect}. If you select such spans, we would like you to (obligatorily):

    \begin{itemize}
        \item preferably give references (e.g., credible websites, academic papers, or books) that show the content is factually wrong, or
        \item give examples that show the content is factually wrong.
    \end{itemize}

\noindent \textbf{Example:}

\noindent\emph{\ul{Question:}} 
Why is it so important for humans to have a balanced nutrition but not for animals? Most animals have a fairly simple diet, carnivores eat only meat their whole life, cows eat exclusively grass etc. So why are human bodies so picky and need a balance of protein, fat, carbs etc from different sources to perform well?

\noindent\emph{\ul{Answer:}} 
\hl{Animals generally have a simpler diet than humans. For example, carnivores only eat meat, while cows only eat grass}...

\noindent\emph{\ul{Reason:}} 
This is a reductionist view of animal nutrition as it doesn't consider how animals have evolved and the complexities of the food chain. For example, lions are carnivores that only eat meat but they eat the stomach of zebras that contain grass/plants and are able to digest it.

\paragraph{3) Relevance.}

You should select a span of text in the \ul{answers} that is \textcolor{red}{irrelevant to answering the question}. Removing such content should not affect the overall quality of an answer. If you select such spans, we would like you to (obligatorily): 

    \begin{itemize}
        \item explain why the selected text is not relevant to answering the question.
    \end{itemize}

\noindent \textbf{Example:}

\noindent\emph{\ul{Question:}} 
What is happening when you get migraines that cause you to lose part of your vision for a short time?

\noindent\emph{\ul{Answer:}} 
\hl{My wife gets these.} An ocular migraine is a condition where the blood vessels in the optic nerve at the back of your eye tighten and swell, resulting in various visual distortions. While classic migraines usually result in intense headaches, sensitivity to light and loud sounds, or nausea, \hl{ocular migraines are not necessarily painful}.

\noindent\emph{\ul{Reason:}} 
Answer contains irrelevant information (writer's wife having them, migraine may not be painful). The person’s wife’s personal health condition doesn’t provide useful information to the question, and the question doesn’t ask about whether ocular migraines are painful or not.

\paragraph{4) Completeness.}

You should: \ul{(a)} select a span of text in the \ul{answer} that does \textcolor{red}{not offer enough details}, or 
\ul{(b)} select \ul{the label Answer 1 or Answer 2} if some \textcolor{red}{relevant information that should be included in the answer is missing}. 
If you select such spans, we would like you to (obligatorily):

\begin{itemize}
    \item offer the details or relevant information that you think should be included. References from credible sources is encouraged. 
\end{itemize}

\noindent \textbf{Example:}

\noindent\emph{\ul{Question:}} 
Why does alcohol make your throat or stomach feel warm when drinking?

\noindent\emph{\ul{Answer:}} 
There are a few reasons why alcohol might make your throat or stomach feel warm. first, alcohol is a solvent, which means it can dissolve other substances. when you drink alcohol, \hl{it can break down the barrier between your stomach and your throat}, causing a burning sensation. second, alcohol can irritate the lining of your stomach, causing inflammation. this can also lead to a burning feeling. finally, when you drink alcohol, your blood vessels dilate, or widen. this increased blood flow can cause a feeling of warmth.

\noindent\emph{\ul{Reason:}} 
The answer should give more details on what the barrier is. [Offer relevant information based on your knowledge, for example, what the barrier is.]

\paragraph{5) References/example/analogy.}

You should select a span of text in the \ul{answer} that is \textcolor{red}{examples/analogies} or a \textcolor{red}{reference to a external credible source} (e.g., websites, papers, or books). If you select such spans, we would like you to (obligatorily):

\begin{itemize}
    \item indicate whether the example/analogy/credible source is helpful, and
    \item give a reason on why it is helpful/not helpful. 
\end{itemize}

\noindent \textbf{Example:}

\noindent\emph{\ul{Question:}} 
How does a speaker vibrate at multiple frequencies simultaneously to deliver sounds to our ears? I'll try and explain what I mean.. say a song is playing - how do the vocals, drums, guitars etc which all have different frequencies (aka a plate is moving back \& forth at a rate to deliver these sound waves) play at the same time? Surely the plate can't vibrate to deliver two or more frequencies at once?

\noindent\emph{\ul{Answer:}} 
\hl{Imagine an ocean with a consistent wave. It flows up and down, with equal distance between the two waves at any time. Now imagine I push a larger, shorter wave into this ocean. The two waves will collide, resulting in some new wave pattern. This new wave pattern is a combination of those two waves.} Speakers work similarly. If I combine two sound waves, I get a new combination wave that sounds different.

\noindent\emph{\ul{Reason:}} 
I like the analogy with the ocean waves, and due to how visual the explanation is it is easier to understand in my opinion.

\paragraph{Answer Preference.}
Based on the five criteria you have went through, please weight all \textcolor{green}{good} and \textcolor{red}{bad} points of each answer and choose the one that is better \textbf{overall}. Writing a motivation to support your choice is obligatory.

\begin{enumerate}
    
\item  When writing your motivation, please refer to the five layers of evaluation. 
\item  If there are other aspects you used to make your decision but are not in the evaluation layers, please mention them in the reason
\item If you have quotations from the answers, please indicate which answer are the quotations from. 
\item Here are some aspects for you to consider (not obligatorily):
    \begin{itemize}
            \item Nice example/analogy, to the point, generic, concise, informative, useful, well structured, easy to follow ...
    \end{itemize}

\end{enumerate}

\paragraph{Overall Requirement.} The overall task requirements are summarized below. Please read them carefully to avoid redoing the task.

\begin{enumerate}

    \item  You have to highlight spans in \textbf{both} question answers for these aspects and \textbf{give reason} why you highlight a span for an aspect.
    \item  Mark as many spans as necessary.
    \item Please be \textbf{objective} in your reasons and avoid using phrases like “I believe” or “I think”.
    \item Your reasons should be informative and succinct.
    \item Please \textbf{use declarative sentences} and avoid using questions in your reasons.
    \item Products like ChatGPT or BARD are absolutely not allowed.

\end{enumerate}

\subsection{Annotation Procedure}
\label{app:sec:anno_procedure}
\begin{figure*}[h]
  \centering
  \includegraphics[width=\textwidth]{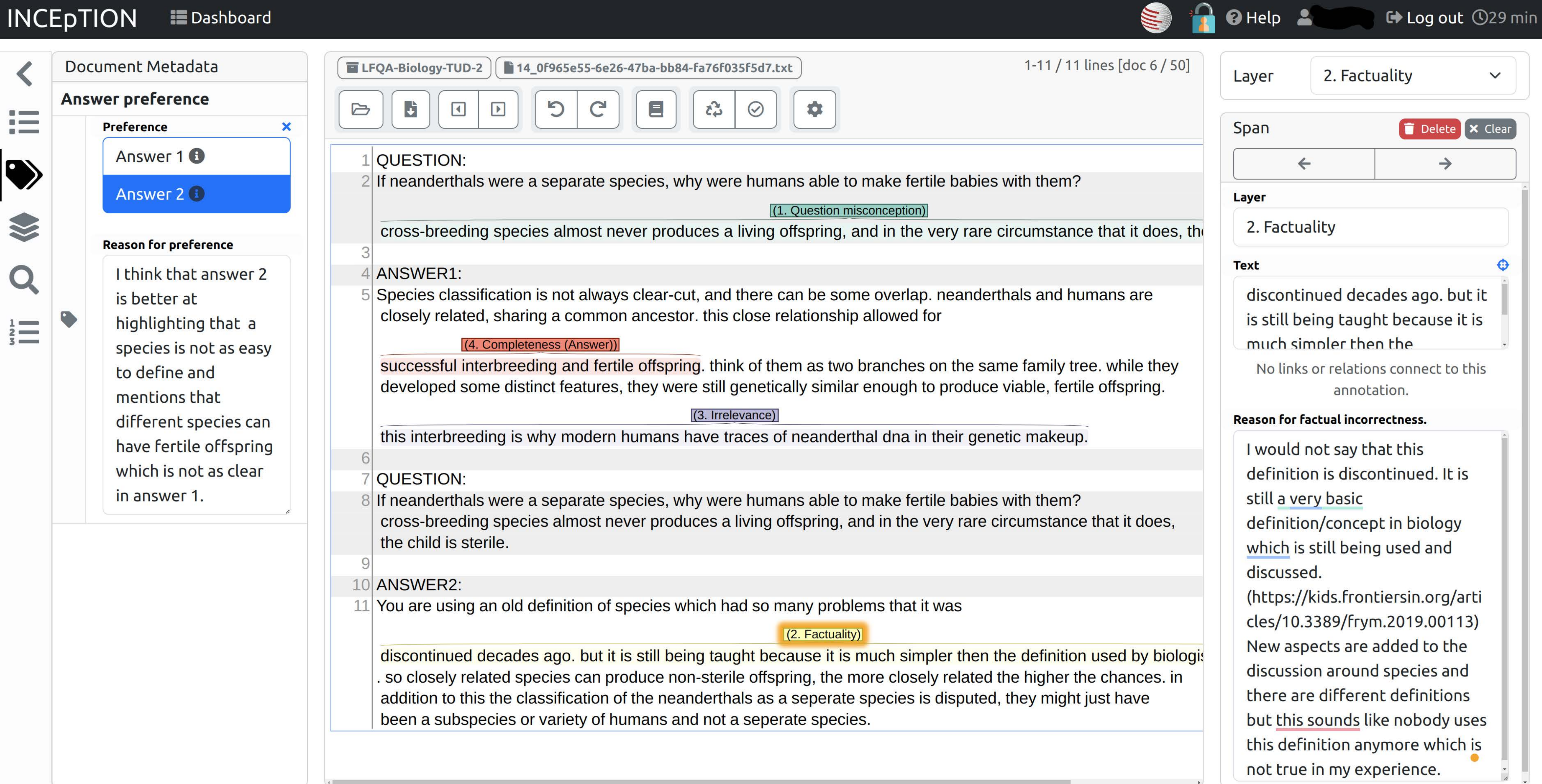}
  \caption{Screenshot of an expert annotated answer on the INCEpTION platform.}
  \label{fig:inception_annotate}
\end{figure*}

The expert annotators spend around 15-20 minutes per question, highlighting the demanding nature of this task. 
We accordingly pay £10/hour and provide a bonus of £10 for good-quality annotations, resulting in a total cost of £3000 to collect expert judgments for 698 questions. The annotators understand that we will use their annotated data for research purposes.
We show a screenshot of an expert annotated answer in \Cref{fig:inception_annotate}.

\newpage
\begin{scriptsize}
\begin{longtable}[h]{p{1.5cm}p{2cm}p{3.5cm}p{3.5cm}p{3.5cm}}
\caption{Additional examples of expert annotated \hl{errors} in long-form answers based on the defined evaluation criteria. The answers in \textcolor{green}{\textit{green}} are expert-preferred answers. The overall answer preference justifications are not shown here.} \\
\label{table:additional_generations} \\
\toprule
\textbf{Aspect} & \textbf{Question} & \textbf{Answer A} & \textbf{Answer B} & \textbf{Expert Justification} \\
\midrule

\multirow{5}{*}{{\parbox{1.8cm}{Question\\Misconception}}}  
& How is GPS free? GPS has made a major impact on our world. \hl{How is it a free service that anyone with a phone can access?} How is it profitable for companies to offer services like navigation without subscription fees or ads?  
& \textit{(human)} GPS "receivers" are just radios that listen to broadcasts by specific satellites, and then use some math to calculate the location. There is no back and forth data transmission at all. So there is no capacity limit and no extra costs for more users. The us government created GPS as a military program. It would likely still be maintained even if no civilians used it, just like it was before it was opened up to the public. 
& \textit{\textcolor{green}{(model)}} GPS is free because it's a public service provided by the US government through a network of satellites. Your phone or device receives signals from these satellites to determine its location. Companies make profits by offering additional services or applications that use GPS data, such as navigational apps or ads based on your location. These companies don't need to pay for access to GPS data because the US government maintains and funds this infrastructure. 
& \begin{minipage}[t]{3.5cm}Although GPS is a free service, made accessible to civilians for free by Ronald Reagan, America's former president at the time, it is not entirely without cost, as U.S. taxpayers fund it. Therefore, if you live and work in the U.S. and use a smartphone, you do not receive GPS on your phone for free, even though you are not directly paying for it; you are paying indirectly through taxes. On the other hand, for people living outside of the U.S., GPS might be available on their smartphones at no additional charge, but they often need to pay indirectly through internet service fees before they can use it. \\Source1: https://www.quora.com/How-is-GPS-sustained-as-a-free-service \\Source2: https://www.gps.gov/policy/funding/ \end{minipage} \\
\midrule

\multirow{4}{*}{{\parbox{1.8cm}{Question\\Misconception}}} 
& If movies run at 24 fps, \hl{why do games look so bad unless they run at at least 60 fps?}  
& \textit{(human)} 
U/dragoniel is right about motion blur, but the more important answer is that movies very rarely move the camera, they certainly don't spin it 180 degrees in a fraction of a second. If they did everything in the shot would be a blurry mess. A whole load of thought goes into how you get the shot you want without too much camera motion and all the bullet time stuff from the matrix was done with many still cameras taking a single photo, not moving a camera on a rocket sled.  
& \textit{\textcolor{green}{(model)}} 
Movies appear smooth at 24 frames per second (fps) due to motion blur created by cameras capturing the natural blur of moving objects. Video games lack this natural motion blur as they consist of sharp, static images. To compensate, games need higher frame rates, like 60 fps, to create smoother transitions between static images, and reduce distractions caused by choppiness or sudden changes while playing. Additionally, games involve frequent interactions with controllers, requiring faster feedback to feel more responsive, which makes higher frame rates even more essential for a good gaming experience.
& \begin{minipage}[t]{3.5cm}This is a misconception. Not all games look bad on FPS below 60. In fact, most games run smoothly between 30 and 60 FPS, although some games will not work smoothly on fps below 60.\\Source: https://www.avg.com/en/signal/9-ways-boost-gaming-rig \end{minipage} \\

\midrule



\multirow{4}{*}{{\parbox{1.8cm}{Factuality}}} 
&  Why can you sleep through the night without going to the bathroom?
& \textit{\textcolor{green}{(human)}}
 Your body produces a hormone called (vasopressin) antidiuretic hormone, or adh, which causes the body to produce less urine at night. It's the reason [...] 
&  \textit{(model)} 
During sleep, your body produces less urine [...] \hl{Additionally, the hormone vasopressin helps to reduce urine production at night.} This combination of decreased [...]
& \begin{minipage}[t]{3.5cm}The urine production is decreased due to the hormone, not additionally. The more concentrated urine is because you cannot drink while sleeping.
\end{minipage} \\ 

\midrule

\multirow{4}{*}{{\parbox{1.8cm}{Factuality}}} 
& Why hasn’t there been anyone to come out as a competitor to ticketmaster?
& \textit{\textcolor{green}{(human)}} 
There are lots of different ticketing companies. Dozens, possibly hundreds. Ticketmaster however is locked in to the biggest venues that feature the biggest artists, either because they have exclusive contracts or they outright own the venue. If you look at smaller venues than stadiums and arenas they often use different ticketing companies.
& \textit{(model)}
Ticketmaster's dominance is due to exclusive contracts with venues and promoters, giving them control over ticket sales. Competitors struggle to break into the market because they lack access to these popular events. \hl{However, some smaller companies like eventbrite and axs have been growing, targeting niche markets and smaller venues to challenge ticketmaster's monopoly.}
& \begin{minipage}[t]{3.5cm}Ticketmaster's market share increased from 66\% to 70\% between 2021 and 2022. Therefore, smaller ticketing companies have so far been unable to challenge Ticketmaster's monopoly.\\Source: https://www.usesignhouse.com/blog/\\ticketmaster-stats \end{minipage} \\

\endfirsthead

\bottomrule
\midrule \multicolumn{5}{r}{{Continued on next page}} \\
\newpage

\toprule
\textbf{Aspect} & \textbf{Question} & \textbf{Answer A} & \textbf{Answer B} & \textbf{Expert Justification} \\
\midrule

\multirow{4}{*}{{\parbox{1.8cm}{Relevance}}} 
& How do publishers of ebooks prevent piracy?
& \textit{(human)}
Some publishers try to only publish on platforms that take complete control of the distribution channels, file types, software and hardware to make it hard for people to copy the ebook without permission. Most famous of these distributors is amazon with their kindle ebook services. But others are doing the same. This does however make it harder for the customers to read the books unless they have the required hardware. And it does not completely prevent piracy. Other publishers make sure that it is easier and better service to buy the book outright instead of pirating it. There is not actually a very big market for pirated ebooks as there is for movies and TV series. \hl{So it can be hard to find pirated versions of the books you want, especially high quality versions. That makes the \$3-10 for an ebook directly from the publisher a preferable deal to a lot of people. The people who do not want to pay this probably would not want to buy the book even if they did not find any pirated versions.} So there is no loss in revenue to not implement any digital rights management.
& \textit{\textcolor{green}{(model)}} 
Ebook publishers use several methods to prevent piracy, which mainly involve digital rights management (DRM) and watermarking. DRM is a technology used to restrict access to and usage of copyrighted material, such as ebooks. In simple terms, DRM is like a lock that prevents unauthorized copying and sharing of ebooks. When you purchase an ebook, you're usually given a key (in the form of an account or device authorization) to unlock and read the book. This prevents users from easily sharing the ebook file with others, as they will not have the necessary key to open it. Some popular drm systems used for ebooks include adobe's adept and amazon's kindle drm. Watermarking is another technique ebook publishers use to deter piracy. Instead of locking the content like drm, a digital watermark embeds identifiable information into the ebook file. This information can be visible (like a logo or text) or invisible (embedded in the file's data). [...]
& \begin{minipage}[t]{3.5cm}This does not tell us how publishers prevent piracy; it only provides a scenario explaining why piracy would not occur in books as it would in visual media. \end{minipage} \\

\midrule

\multirow{4}{*}{{\parbox{1.8cm}{Relevance}}} 
& Why do companies like Meta have so many employees, what do they do?
& \textit{(human)}
In my experience software companies lean heavily on product and project managers. To stay competitive in tech you have to constantly be trying to innovate (at least that's the idea), which takes a lot of people to come up with and plan out these projects, even if many never see the light of day. Beyond that, maintaining a tech infrastructure like that is very complex,, goes well beyond devs. [...] Facebook also has a huge footprint in the open source community, in that they actively maintain many sophisticated projects they built for their products that are now used by millions of other products, like their react frameworks. Not to mention they heavily use other open source softwares and dedicate some resources to maintaining those as well [...] \hl{Successful tech companies are incredibly convoluted and complex in my experience. Many times there are whole teams working on "nothing" because their part of the business is on its way out, or they were part of another company that was acquired, and in either case, someone in the middle hasn't gotten around to trimming the fat/reallocating those resources. Source: have worked for a faang company, other huge tech companies, now work as a private contractor bc of how ** annoying these conglomerates are to work with.}
& \textit{\textcolor{green}{(model)}} 
Meta, formerly known as Facebook, is a massive tech company that requires a large number of employees to manage and maintain its various services, products, and operations. [...] let's break it down into main areas:1. Software development: Meta operates numerous services and applications, including Facebook, Instagram, Whatsapp, and the oculus virtual reality platform. Each of these services requires teams of software engineers, product managers, and designers to develop, maintain, and improve the platforms. [...] 2. Infrastructure and operations: with billions of active users, Meta must maintain and develop a vast infrastructure of servers, networks, and data centers to handle the enormous amount of data being generated, processed, and stored. [...] 3. Research and development: companies like meta are always innovating and creating new technologies. They require extensive r\&d teams that include experts in fields like artificial intelligence, virtual reality, and machine learning. These teams drive the development of cutting-edge technologies and solutions for Meta's products. [...] 4. Sales and marketing: meta generates a significant portion of its revenue from advertising. [...] 5. User support and moderation: with billions of users worldwide, [...]
& \begin{minipage}[t]{3.5cm}Adds nothing to the explanation is a rant by the person. \end{minipage} \\ 

\bottomrule
\midrule \multicolumn{5}{r}{{Continued on next page}} \\
\newpage

\toprule
\textbf{Aspect} & \textbf{Question} & \textbf{Answer A} & \textbf{Answer B} & \textbf{Expert Justification} \\
\midrule

\multirow{5}{*}{{\parbox{1.8cm}{Completeness}}}  
& How do pharmacies work? Do they just have every kind of medication at all time? How is a prescription ready within an hour? Eta thank you everyone for taking the time to respond! I know it seems super obvious as to how it works, this was a late night thought I had and needed to know?
& \textit{(human)} General yes, pharmacies will keep enough of the more commonly requested medications in stock so that most prescriptions can be filled the same day. \hl{When that's not possible, depending on various factors,} many pharmacies may have multiple deliveries from suppliers each week, making it possible to order anything that isn't in stock within a few days if it's available from the supplier.
& \textit{\textcolor{green}{(model)}} Pharmacies keep a stock of commonly prescribed medications on hand. When you provide your prescription, the pharmacist checks their inventory, measures and packages the drug, and puts the proper labeling and instructions on it. If the medication isn't in stock, they can usually order it and have it ready in a short time. Efficient organization and well-trained staff help facilitate the quick turnaround for prescriptions.
& \begin{minipage}[t]{\linewidth} The answer should specify factors that may affect the availability of medications, such as manufacturing issues, disruptions in supply, and stockouts of essential medicines (World Health Organization). For more information, you can visit: https://www.who.int/our-work/access-to-medicines-and-health-products \end{minipage} \\

\midrule

\multirow{5}{*}{{\parbox{1.8cm}{Completeness}}}  
& What are the pros/cons of borrowing money from the bank vs paying cash when buying a house? And why would it ever be a good idea to borrow if you can pay cash and pay no interest to the bank?
& \textit{\textcolor{green}{(human)}}
If you can make more money doing something else with your cash than using it to pay down your mortgage, then it's beneficial to take out a loan. It depends on your credit rating what kind of interest you'll get on your loan, and what the long term stock market trends are, or if you want to buy multiple properties at the same time, for example. But making money in the stock market is not a given, so there is risk involved in going that route. Also, you want to have cash on hand for possible emergency situations, like big unexpected medical bills, your car dies, you lose your job, etc. On the other hand, if you lose your job and you have a mortgage, then you might end up defaulting on your loan, and lose the home and any of the equity you had in it.
& \textit{(model)}
\hl{The main advantage of borrowing money from a bank (a mortgage) to buy a house is that it allows you to purchase a property without having all the cash up front. This enables people to become homeowners earlier in life and potentially benefit from the appreciation of the home's value. Paying a mortgage also helps build credit history, which can be beneficial for obtaining future loans or credit cards. On the other hand, paying cash for a house has the advantage of no interest costs, which can save a substantial amount of money over time. Cash buyers may also have more negotiating power, as they are not reliant on bank approval and can close the deal quickly. In certain situations, it can be better to borrow money even if you have the cash. This is because investing the cash into diversified investments (stocks, bonds, etc. ) may provide better returns than paying off the mortgage early, although it comes with some level of risk. Additionally, having cash reserves can provide financial flexibility for emergencies or future opportunities.}
& \begin{minipage}[t]{\linewidth} This answer is missing the disadvantages of getting a mortgage and paying in cash. It could mention how getting a mortgage means you will have to pay interest, as well as any mortgage fees (such as arrangement fees or valuation fees). It could also discuss how your home could be put at risk if you are unable to keep up with the repayments since a mortgage is secured against the value of your property. In terms of paying in cash, the answer could explain that this will narrow your investment portfolio, thereby missing the opportunity to invest in other assets that could increase your wealth over time. \\Source: (Heming,2022) https://www.moneysupermarket.com/\\mortgages/advantages-and-disadvantages/ \end{minipage} \\

\midrule

\multirow{5}{*}{{\parbox{1.8cm}{References}}}  
& Why can’t spam call centers be automatically shut down? Additionally, why can’t spam calls be automatically blocked, and why is nobody really doing a whole lot about it? It seems like this is a problem that they would have come up with a solution for by now.
& \textit{(human)} Most of the spam call centers originate in countries where the governments don't \_really\_ care all that much. \hl{For example, a good number of them are in smaller cities in India, where the politicians are willing to turn a blind eye for the right price (and so long as they don't go around scamming Indians). From there, the call centers are just another overseas call.} There are a number of \_legitimate\_ call centers in foreign countries, so looking to block the scam ones is a needle in a stack of needles. Its a ton of effort for the phone company with little actual benefit to them.
& \textit{\textcolor{green}{(model)}} Spam call centers can't be automatically shut down because they are often located in countries with weak regulation and they use advanced techniques to hide their identity, like caller id spoofing. Blocking spam calls is challenging because these callers use ever-changing phone numbers, making it difficult to maintain an updated block list. While there are some efforts to combat spam calls, such as telecom companies implementing call-filtering systems and governments enacting regulations, the issue is deeply rooted in the economic incentive for spam callers and the complexities in tracking them down. Ultimately, tackling spam calls requires international cooperation and continuous technology advancements.
& \begin{minipage}[t]{\linewidth}The example is not helpful in addressing the query. It is a controversial example that will stir up the emotions of readers, and therefore distract them. \end{minipage} \\
\bottomrule

\end{longtable}
\end{scriptsize}


\begin{scriptsize}
\begin{longtable}[h]{p{2cm}p{4cm}p{4cm}p{4cm}}
\caption{
Examples of answers refined using our Error-informed Refinement approach. We show the feedback from the error-feedback model and the corresponding refined answers from the refinement model incorporating the feedback. The highlighted portion in the refined answer depicts the feedback incorporated in the answer.
} \\
\label{table:eir_examples} \\
\toprule
 \textbf{Question} & \textbf{Answer} & \textbf{Feedback} & \textbf{Refined Answer} \\
\midrule

[ELI5] Why do cars sound different in reverse?
& Reverse gears often are straight-cut instead of cross-cut because they get used far less and are cheaper to make.[...] The gears on the top are cross-cut and make far less noise and last a lot longer. Straight cut makes more noise because the teeth sort of slap against each other as the gears mesh. Cross cut are quiet because the more gently mesh with each other over a diagonal.
& \begin{minipage}[t]{4cm} 
Consistency score: 1.0\\
Prediction:\\
1. [Incomplete] Reasons: This is correct but could be improved by mentioning that straight-cut gears are louder because there is more friction between the teeth when they meet. This causes more heat buildup and wear on the teeth, which contributes to the noise. [...]
\\2. [Complete]
\\3. [Complete]
\\4. [Complete]
\\5. [Complete]
\end{minipage}

& \begin{minipage}[t]{4cm}
Cars can sound different in reverse due to the type of gears used in the transmission. Reverse gears are often straight-cut instead of cross-cut because they are used less frequently and are therefore cheaper to manufacture. \hl{Straight-cut gears have teeth that meet end-to-end, causing more friction and heat buildup, which results in a louder noise.} In contrast, cross-cut gears have teeth that mesh at an angle, reducing the amount of friction and heat generated, resulting in a quieter noise. Additionally, straight-cut gears tend to have more play between the teeth, which can also contribute to the noise. To hear the difference for yourself, you can listen to a video of a car reversing with both straight-cut and cross-cut gears.[...]
\end{minipage} \\
\midrule

[ELI5] How have they brought air inside the International Space Station? In liquid form?
& In liquid form? In a way, yeah. Usually in the form of water, which is safe to transport, doesn't have to be pressurized and you only have to vent the excess hydrogen.
& \begin{minipage}[t]{4cm} 
Consistency score: 1.0\\
Prediction:\\
1. [Incomplete] Reasons: The answer does not completely describe how this is possible. It mentions that water is used as a source of oxygen, but it fails to explain how this works. Additionally, it does not address the issue of bringing air into the ISS from an external source.
\\2. [Complete]
\\3. [Complete]

\end{minipage}

& \begin{minipage}[t]{4cm}
To bring air inside the International Space Station (ISS), the astronauts use a process called \"oxygen generation,\" where they convert carbon dioxide and other gases present in the space station's atmosphere into breathable oxygen. This process involves several steps:\hl{\\1. Carbon Dioxide Removal: [...]\\2. Oxygen Generation: [...] \\3. Air Revitalization: [...]\\4. Humidification: [...]\\5. Filtering: [...]\\In terms of bringing air into the ISS from an external source, there are actually two methods used:\\1. Atmospheric Replenishment: [...]\\2. Cargo Vehicle Supply: [...] }\\So, to summarize, the ISS has systems in place to generate oxygen from carbon dioxide and other gases present in the space station's atmosphere, as well as methods for bringing in air from external sources when necessary.
\end{minipage} \\
\midrule

[ELI5] Why are chairs slightly incurved instead of the other way around?
& Some busses and trains in Sweden actually have chairs that make you sit with a straight back, but i think It's cause people find those types of chairs more comfortable.
& \begin{minipage}[t]{4cm} 
Consistency score: 1.0\\
Prediction:\\
1. [Incomplete] Reasons: The answer does not explain why most chairs are slightly inclined.

\end{minipage}

& \begin{minipage}[t]{4cm}
Chairs are designed with a slight incline rather than being perfectly straight for several reasons. \hl{Firstly, an angled seat allows users to maintain better posture while sitting. When the backrest is at a slight angle, it supports the natural curvature of the spine, promoting good posture and reducing strain on the back and neck muscles. This can help prevent long-term health issues associated with poor sitting positions, such as lower back pain and herniated discs.\\Additionally, the inclination of the chair helps to distribute weight evenly across the seat and backrest, reducing pressure points and providing a more comfortable sitting experience. This is particularly important for people who spend extended periods of time sitting, such as office workers or students.}\\While some buses and trains in Sweden may have straight-backed seats, this design is not typical of most chairs found in everyday use. In fact, many ergonomic experts recommend that chairs be designed with a slight incline to promote proper posture and reduce discomfort.[...]
\end{minipage} \\

\endfirsthead

\bottomrule
\midrule \multicolumn{4}{r}{{Continued on next page}} \\
\newpage

\toprule
\textbf{Question} & \textbf{Answer} & \textbf{Feedback} & \textbf{Refined Answer} \\
\midrule

[ASQA] When did the first cellular phone come out?
& The first handheld cellular mobile phone was demonstrated by John F. Mitchell and Martin Cooper of Motorola in 1973. In 1983, the DynaTAC 8000x was the first commercially available handheld mobile phone. The phone had a talk time of just thirty minutes and took ten hours to charge.
& \begin{minipage}[t]{4cm} 
Consistency score: 0.55\\
Prediction:\\
1. [Incomplete] Reasons: This is correct but it would be useful to mention that this demonstration didn't lead to the development of an actual product. Instead, it showed the potential of such a device.\\
2. [Complete]\\
3. [Complete]\\

\end{minipage}

& \begin{minipage}[t]{4cm}
The first handheld cellular mobile phone was demonstrated by John F. Mitchell and Martin Cooper of Motorola in 1973. \hl{This demonstration showcased the potential of a portable device that could communicate wirelessly over long distances, revolutionizing the way people communicated while on-the-go. However, it wasn't until 1983 that the first commercially available handheld mobile phone, the DynaTAC 8000x, became available to consumers.} With a talk time of just thirty minutes and a charging time of ten hours, this early model paved the way for future advancements in mobile technology. Today, smartphones have become an essential part of modern life, offering a wide range of features and capabilities beyond simply making calls and sending texts.
\end{minipage} \\

\midrule

[ASQA] Where is the danube river located in europe?
& The Danube river is the second longest river in Europe, running through Central and Eastern Europe. Starting in Germany the river flows through Austria, Slovakia, Hungary, Croatia, Serbia, Romania, Bulgaria, Moldova and Ukraine before draining into the Black Sea. The river flows through many cities including Pančevo, Drobeta-Turnu Severin, Bratislava, Ruse, Novi Sad, Ulm, Vienna, Zemun, Vidin, Regensburg, Ingolstadt, Linz, Győr, Dunaújváros, Smederevo, Izmail, Belgrade, Apatin, Galați, Vukovar and Budapest.

& \begin{minipage}[t]{4cm}
Consistency score: 0.70\\
Prediction:\\
1. [Complete]\\
2. [Complete]\\
3. [Incomplete] Reasons: These are just names of cities; it would be useful to know where they are located (which countries) so you can understand how the river flows through different nations.

\end{minipage}

& \begin{minipage}[t]{4cm}
The Danube River is the second longest river in Europe, flowing through Central and Eastern Europe for a total length of approximately 2,857 kilometers (1,777 miles). It originates in the Black Forest region of Germany, then passes through or along the borders of several other European countries, including Austria, Slovakia, Hungary, Croatia, Serbia, Romania, Bulgaria, Moldova, and Ukraine before emptying into the Black Sea.

Along its journey, the Danube River flows through numerous cities and towns, some of which include:\\

* \hl{In Germany: Ulm, Ingolstadt, and Regensburg}
* \hl{In Austria: Linz and Vienna}
* \hl{In Slovakia: Bratislava}
* \hl{In Hungary: Budapest, Gyor, and Szeged}
* \hl{In Croatia: Vukovar and Osijek}
* \hl{In Serbia: Belgrade and Smederevo}
* \hl{In Romania: Bucharest and Galati}
* \hl{In Bulgaria: Ruse and Vidin}
* \hl{In Moldova: Kishinev}
* \hl{In Ukraine: Odessa and Reni} \\

Each of these cities has a unique history and culture, [...].
\end{minipage} \\

\midrule

[HQ$^2$A] How did human trafficking become a thing? Why do people kidnap and sell others?
& Human trafficking became a thing likely in pre-history and hasn't stopped at all. There is always a need for people to do something and being sent from one place to another. Various forms of slavery is obviously a prominent activity and has been present in many societies since pre-history and still exists today.
& \begin{minipage}[t]{4cm} 
Consistency score: 0.55\\
Prediction:\\
1. [Incomplete] Reasons: This is correct; it mentions that human trafficking has occurred since prehistoric times and continues today. However, it would be useful to trace the historical emergence of modern slave trade and human trafficking. For example, it could explain how during the transatlantic slave trade, African slaves were transported across the Atlantic Ocean to the Americas \\(https://www.britannica.com/topic/\\transatlantic-slave-trade). This would help to provide context and give further insight into the history of human trafficking.\\
2. [Complete]\\
3. [Complete]\\

\end{minipage}

& \begin{minipage}[t]{4cm}
Human trafficking, unfortunately, has a long and complex history that dates back to pre-historic times. The practice of capturing and enslaving individuals for various purposes, including forced labor and sexual exploitation, can be traced back to ancient civilizations such as Greece, Rome, and Egypt. \hl{However, the modern form of human trafficking, which involves the use of force, fraud, or coercion to exploit people for labor or commercial sex, gained momentum during the transatlantic slave trade era.}\\\hl{During this period, millions of Africans were forcibly taken from their homes and sold into slavery in the Americas, where they were subjected to brutal working conditions and treated as property rather than human beings. This tragic chapter in human history laid the foundation for the modern human trafficking industry, which has evolved over time to include a wide range of activities, such as child labor, domestic servitude, [...]}\\So, why do people engage in human trafficking? The reasons are complex and varied, but some common motivations include financial gain, power and control, and a lack of empathy for the victims. Traffickers often target vulnerable individuals, such as those living in poverty or fleeing conflict, and use manipulation, coercion, and violence to keep them [...]

\end{minipage} \\ 

\bottomrule
\midrule \multicolumn{4}{r}{{Continued on next page}} \\
\newpage

\toprule
\textbf{Question} & \textbf{Answer} & \textbf{Feedback} & \textbf{Refined Answer} \\
\midrule

[HQ$^2$A] How are e-fits constructed in the pursuit of a suspect?
& E-fits, or electronic facial composites, are digital images created to help identify suspects in criminal investigations. They are constructed by combining different facial features based on the eyewitness's description of the suspect. The process starts with an interview between the eyewitness and a trained police officer or forensic artist. The officer uses specialized software, which contains a database of various facial features, such as eyes, noses, mouths, and hairstyles. The eyewitness selects the features that best match their memory of the suspect's appearance. Once the initial facial composite is created, the officer and the eyewitness work together to refine the image. They may adjust the size, shape, and position of features to improve accuracy. The final e-fit is then used to aid police investigations, typically by sharing with the public and other law enforcement agencies, in the hopes of identifying the suspect.

& \begin{minipage}[t]{4cm} 
Consistency score: 1.0\\
Prediction:\\
1. [Complete]\\
2. [Complete]\\
3. [Complete]\\
4. [Complete]\\
5. [Complete]\\
6. [Complete]\\
7. [Complete]\\
8. [Incomplete] Reasons: It would be helpful if this section elaborated more on how e-fits are shared and why it is beneficial to do so.

\end{minipage}

& \begin{minipage}[t]{4cm}
E-fits, or electronic facial composites, are digital images created to help identify suspects in criminal investigations. They are constructed by combining different facial features based on the eyewitness's description of the suspect. The process starts with an interview between the eyewitness and a trained police officer or forensic artist. The officer uses specialized software, which contains a database of various facial features, such as eyes, noses, mouths, and hairstyles. The eyewitness selects the features that best match their memory of the suspect's appearance. Once the initial facial composite is created, the officer and the eyewitness work together to refine the image. They may adjust the size, shape, and position of features to improve accuracy.\\\hl{Once the e-fit is complete, it can be shared with the public and other law enforcement agencies through various channels, such as social media, local news outlets, and online crime reporting platforms.} Sharing the e-fit helps to increase the chances of identifying the suspect, as it allows a larger pool of people to view and recognize the individual. Additionally, sharing the e-fit can also serve as a deterrent to the suspect, as they may become aware that the authorities are actively searching for them.\\\hl{In addition to being shared with the public, e-fits can also be distributed to other law enforcement agencies, such as federal authorities, neighboring jurisdictions, and international partners. This helps to expand the search area and increase the likelihood of identifying the suspect.} Furthermore, e-fits can be updated and revised as new information becomes available, allowing the investigation to continue until the suspect is apprehended. [...]

\end{minipage} \\

\bottomrule

\end{longtable}
\end{scriptsize}